\documentclass[manuscript,screen]{acmart}
\AtBeginDocument{%
  }

\setcopyright{acmlicensed}
\copyrightyear{2018}
\acmYear{2018}
\acmDOI{XXXXXXX.XXXXXXX}

\acmJournal{JACM}
\acmVolume{37}
\acmNumber{4}
\acmArticle{111}
\acmMonth{8}

\usepackage{seqsplit}
\usepackage{enumitem}
\usepackage{tablefootnote}
\usepackage{tabulary}
\usepackage{colortbl}
\usepackage{algorithm}
\usepackage{algpseudocode}
\usepackage{textcomp}
\usepackage{amsmath}
\usepackage{xcolor}
\usepackage{calc}
\usepackage{subcaption}
\usepackage{wrapfig}
\usepackage{threeparttable}
\usepackage{multirow}
\usepackage[skins]{tcolorbox}
\usepackage{algpseudocode}
\usepackage{adjustbox}



\usepackage[framemethod=tikz]{mdframed}
\mdfdefinestyle{customstyle}{
  linecolor=gray!100,
  linewidth=3pt,
  innerleftmargin=3pt,
  topline=false,
  rightline=false,
  bottomline=false,
  leftline=true,
  innerrightmargin=3pt,
  innertopmargin=3pt,
  innerbottommargin=3pt,
  backgroundcolor=gray!15
}
\newenvironment{custommdframed}
  {\begin{mdframed}[style=customstyle]}
  {\end{mdframed}}

\mdfdefinestyle{noleftline}{
  linecolor=gray!100,
  linewidth=3pt,
  topline=false,
  rightline=false,
  bottomline=false,
  leftline=false, 
  innerleftmargin=3pt,
  innerrightmargin=3pt,
  innertopmargin=3pt,
  innerbottommargin=3pt,
  backgroundcolor=gray!15,
}
\newenvironment{noleftmdframed}
  {\begin{mdframed}[style=noleftline]}
  {\end{mdframed}}

\begin{document}

\title{LLM-Driven Collaborative Model for Untangling Commits via Explicit and Implicit Dependency Reasoning}

\author{Bo Hou}
\orcid{0009-0003-4743-9432}
\affiliation{%
  \institution{State Key Laboratory of Complex \& Critical Software Environment (SKLCCSE), School of Computer Science and Engineering, Beihang University}
  \city{Beijing}
  \country{China}
}
\email{houbo2024@buaa.edu.cn}

\author{Xin Tan}
\orcid{0000-0003-1099-3336}
\authornote{Corresponding author.}
\affiliation{%
  \institution{SKLCCSE, School of Computer Science and Engineering, Beihang University}
  \city{Beijing}
  \country{China}
}
\email{xintan@buaa.edu.cn}

\author{Kai Zheng}
\orcid{0009-0008-3707-2477}
\affiliation{%
  \institution{SKLCCSE, School of Computer Science and Engineering, Beihang University}
  \city{Beijing}
  \country{China}
}
\email{21373443@buaa.edu.cn}

\author{Fang Liu}
\orcid{0000-0002-3905-8133}
\affiliation{%
  \institution{SKLCCSE, School of Computer Science and Engineering, Beihang University}
  \city{Beijing}
  \country{China}
}
\email{fangliu@buaa.edu.cn}

\author{Yinghao Zhu}
\orcid{0000-0002-2640-6477}
\affiliation{%
  \institution{School of Computing and Data Science, The University of Hong Kong}
  \city{Hong Kong SAR}
  \country{China}
}
\email{zhuyinghao@buaa.edu.cn}

\author{Li Zhang}
\orcid{0000-0002-2258-5893}
\affiliation{%
  \institution{SKLCCSE, School of Computer Science and Engineering, Beihang University}
  \city{Beijing}
  \country{China}
}
\email{lily@buaa.edu.cn}

\renewcommand{\shortauthors}{Hou et al.}

\begin{abstract}
Atomic commits, which address a single development concern, are a best practice in software development. In practice, however, developers often produce tangled commits that mix unrelated changes, complicating code review and maintenance. Prior untangling approaches—rule-based, feature-based, or graph-based—have made progress but typically rely on shallow signals and struggle to distinguish explicit dependencies (e.g., control/data flow) from implicit ones (e.g., semantic or conceptual relationships).
In this paper, we propose \textit{ColaUntangle}, a new collaborative consultation framework for commit untangling that models both explicit and implicit dependencies among code changes. \textit{ColaUntangle} integrates Large Language Model (LLM)-driven agents in a multi-agent architecture: one agent specializes in explicit dependencies, another in implicit ones, and a reviewer agent synthesizes their perspectives through iterative consultation. To capture structural and contextual information, we construct Explicit and Implicit Contexts, enabling agents to reason over code relationships with both symbolic and semantic depth.
We evaluate \textit{ColaUntangle} on two widely-used datasets (1,612 C\# and 14k Java tangled commits). Experimental results show that \textit{ColaUntangle} outperforms the best-performing baseline, achieving an improvement of 44\% on the C\# dataset and 82\% on the Java dataset. These findings highlight the potential of LLM-based collaborative frameworks for advancing automated commit untangling tasks.
\end{abstract}



\received{20 February 2007}
\received[revised]{12 March 2009}
\received[accepted]{5 June 2009}

\maketitle

\section{Introduction}

In collaborative software development, a \textit{commit} is the basic unit of code change, consisting of source file modifications and a message summarizing their purpose \cite{shen2021smartcommit, dong2022fira, gomez2015visually}. Ideally, each commit should be \textit{atomic}, addressing a single concern—such as implementing a feature, fixing a bug, or refactoring \cite{li2022utango,guo2019decomposing,ram2018makes,robillard2002concern}. Atomic commits are a widely endorsed best practice for improving readability, facilitating reviews, and helping maintenance by clearly linking changes to their intention \cite{tao2012empiricalwork,rigby2012contemporary,wang2019cora,herbold2022fine,baysal2016investigating}. As a result, many software communities and companies recommend small, single-purpose commits \cite{fan2024HD-GNN,linuxSubmittingPatches,microsoftGitGuidance}.

However, in practice, developers often create \textit{tangled commits}, which combine code changes related to multiple development concerns within a single commit \cite{fan2024HD-GNN,murphy2011we}. \textit{Tangled commits} commonly arise due to time constraints or unclear boundaries between concerns during implementation \cite{fan2024HD-GNN,parțachi2020flexeme,li2022utango}. For example, a developer may simultaneously refactor code while fixing a bug alongside documentation updates, resulting in a commit that does not adhere to the principle of separation of concerns \cite{herzig2016impact,herzig2013impact,nguyen2013filtering}.

Empirical studies \cite{herzig2013impact,herzig2016impact,tao2015partitioning,nguyen2013filtering} find that \textit{tangled commits} are prevalent, indicating that 11\% to 40\% of commits in software repositories are tangled. Such commits hinder program comprehension, complicate maintenance and reduce code quality~\cite{tao2012empiricalwork,parțachi2020flexeme,li2015gated}. 
Moreover, \textit{tangled commits} reduce the accuracy of automated tools such as bug prediction and localization models that rely on mining repository history \cite{herzig2013impact,herzig2016impact,zhou2019devign,kirinuki2014hey,zimmermann2005mining,guo2017interactively}. These tools often assume that each commit addresses a single concern, so \textit{tangled commits} introduce noise and degrade model performance.

Researchers have proposed various approaches to untangling commits to split a \textit{tangled commit} into separate coherent changes. These methods fall into three categories based on their automation level: \textit{heuristic rule-based}, \textit{feature-based}, and \textit{graph clustering-based}. Heuristic methods use manually defined rules—such as line distance, textual similarity, or co-change frequency—to assess change relatedness~\cite{barnett2015Heuristic-rule,herzig2013impact,kirinuki2016splitting,wang2019cora}. Feature-based methods employ handcrafted features (e.g., same method, class, or package) in supervised models to classify change pairs~\cite{dias2015EpiceaUntangler,yamashita2020changebeadsthreader}. Graph clustering-based methods build program graphs, e.g., program dependency graphs (PDGs), to model structural dependencies and use clustering or node embeddings to group related changes~\cite{parțachi2020flexeme,li2022utango,xu2025COMUNT,fan2024HD-GNN}.

Despite their methodological differences, these approaches share several key limitations. First, they often rely on surface-level signals or structural proximity and lack the capacity for deeper semantic reasoning. Graph-based methods attempt to encode some semantic relationships via graph embeddings, but they typically require substantial training and are limited in their ability to generalize across projects or languages. Second, most approaches act as black-box models, offering limited interpretability and little insight into why specific code changes are grouped together. Third and most critically, prior work does not clearly distinguish or integrate different types of dependencies among code changes. For instance, two edits may be connected via control flow (explicit) or may reflect a coherent conceptual refactoring (implicit), yet existing models either fail to recognize such links or treat them uniformly, resulting in reduced effectiveness on complex or ambiguous commits.

We argue that effective commit untangling must go beyond surface-level patterns and rely on a principled framework that explicitly accounts for both \textit{explicit} and \textit{implicit} dependencies. \textit{Explicit dependencies} refer to observable relationships such as control or data flow, containment, or static code references. \textit{Implicit dependencies} involve semantic connections—such as conceptual similarity, logical association, or shared intent—that are not necessarily reflected in the program structure. While some existing methods may partially capture these relationships through learned embeddings or co-change patterns, they do not explicitly model or directly leverage these dependencies as integral elements of the untangling process.

Recent advances in large language models (LLMs) have demonstrated strong performance in software engineering tasks that require semantic understanding, reasoning, and explanation~\cite{yang2024harnessing,fan2023large,hou2024large,fan2025exploring,li2024understanding}. LLMs are particularly well-suited to identify implicit dependencies that go beyond what is detectable via static analysis or statistical patterns. However, a single LLM acting alone may not consistently balance different dependency perspectives or resolve ambiguous cases. Inspired by recent work on multi-agent collaboration~\cite{he2025llm, wang2025colacare, tang2023medagents, Kim2024mdagents}, we propose decomposing the untangling task into subtasks, each handled by specialized agents, and enabling them to collaborate through consultation.

To this end, we introduce \textit{ColaUntangle}, a collaborative consultation framework for commit untangling that integrates both explicit and implicit dependencies. We construct \textit{Explicit and Implicit Contexts} to capture explicit and implicit dependencies between code changes.
Then, we design a multi-agent architecture driven by LLMs: an explicit worker agent focused on explicit dependencies (e.g., control/data flow), an implicit worker agent focused on implicit dependencies (e.g., conceptual relationships), and a reviewer agent that synthesizes and reconciles their results. All agents provide not only untangling decisions but also explanations. Through iterative interaction, these agents simulate human-like consultation and collectively decide on the final untangling outcome.

To evaluate the effectiveness and efficiency of the proposed approach, we conduct experiments on the widely-used C\# and Java datasets~\cite{parțachi2020flexeme,li2022utango,fan2024HD-GNN} with 1,612 and 14k \textit{tangled commits}. The evaluation results show that \textit{ColaUntangle} achieves better untangling results (i.e., 44\% and 82\% enhancement of effectiveness for C\# and Java, compared to the best-performing baseline). Overall, our work makes the following contributions:
\begin{itemize}
    \item \textbf{\textit{ColaUntangle}: the first collaborative consultation model leveraging LLM-driven agents for commit untangling.} We propose \textit{ColaUntangle}, the first LLM-driven collaborative consultation model that untangles commits by considering explicit and implicit dependencies among code changes.
    \item \textbf{Formal and interpretable evidence framework.} We define a clear and generalizable set of Explicit and Implicit Dependencies between code changes. This principled evidence framework provides direct guidance for LLM reasoning, which is foundational for explainable untangling.
    \item \textbf{A unified LLM framework for explainable, human-aligned untangling.} We introduce a unified framework leveraging LLM-driven reasoning to infer semantics, robustly balance explicit and implicit dependencies, identify cosmetic edits and generate contextual explanations. Unlike prior black-box untangling models, this collaborative and human-centric architecture achieves explainable and interpretable untangling results.
    \item \textbf{Extensive empirical evaluation.} We evaluate \textit{ColaUntangle} against the state-of-the-art approaches, and the results demonstrate its superior performance. Beyond quantitative results, we conduct a detailed examination of both correctly untangled and misclassified cases, revealing how explainable reasoning contributes to accurate decisions and exposing the boundaries of current LLM-based untangling. 
\end{itemize}

\section{Related Work} 

Early research on commit untangling primarily relied on heuristic rules. Barnett et al.~\cite{barnett2015Heuristic-rule} introduced the concept of diff-regions and proposed CLUSTERCHANGES to address the lack of automated support. Building on this, Herzig et al.~\cite{herzig2016impact} extracted relationships between code changes using the Confidence Voter and applied clustering algorithms to group diff-hunks. However, these approaches predominantly depended on static analysis techniques, limiting their applicability to dynamically typed languages.

\textcolor{black}{
Subsequent efforts introduced feature-based classification methods. Dias et al.~\cite{dias2015EpiceaUntangler} developed EpiceaUntangler, which mines relationships among code changes and employs a random forest classifier. This approach was suitable for untyped language environments where static analysis is inherently constrained. Later, ChangeBeadsThreader~\cite{yamashita2020changebeadsthreader}, a streamlined variant of EpiceaUntangler, enabled developers to interactively refine clustering results.
}

\textcolor{black}{
More recent studies have shifted toward graph clustering combined with machine learning. Flexeme~\cite{parțachi2020flexeme} introduced Multi-version Program Dependency Graph ($\delta$-PDG) and Multi-version Name Flow Graph ($\delta$-NFG) and applied hierarchical clustering algorithms to untangle commits. SmartCommit~\cite{shen2021smartcommit} presented an interactive graph partitioning approach utilizing a scalable Diff Hunk Graph representation. It employed an edge shrinking algorithm to partition the graph into subgraphs and enabled developers to adjust outcomes.
}

\textcolor{black}{
Graph neural networks (GNNs) \cite{li2015gated,xu2018powerful,kipf2016semi} have also been applied to untangling tasks. UTango~\cite{li2022utango} was the first method to incorporate GNNs, reusing Flexeme’s $\delta$-PDG while enriching it with node context information. By integrating clone detection, UTango overcame Flexeme’s limitations in handling implicit dependencies among cloned code. HD-GNN~\cite{fan2024HD-GNN} further advanced this line of work by proposing a fine-grained hierarchical graph model that captures the global context of code changes at both the entity and statement levels, addressing the problem of neglected hidden dependencies in prior approaches. More recently, Xu et al.~\cite{xu2025COMUNT} proposed an attributed graph modeling approach for detecting and untangling composite commits. Their method constructed a graph of code statements and dependencies, leveraging graph neural networks and clustering algorithms to achieve untangling. 
}

\section{Motivation}
High-quality software development demands atomic commits, where each commit addresses a single, isolated concern. However, in practice, developers frequently introduce tangled commits containing changes related to multiple independent activities. While prior work has attempted to untangle these commits, they often fall short due to an inability to fully capture the diverse and often subtle relationships between code changes. To define the scope of the problem and motivate the design of \textit{ColaUntangle}, we analyze real-world commit examples, which crystallize four critical challenges faced by existing untangling methods.

\subsection{Motivating Examples and Core Challenges}
\label{challenges}
\begin{figure*}
    \centering
    \setlength{\abovecaptionskip}{0.1cm}
   \includegraphics[width=\linewidth]{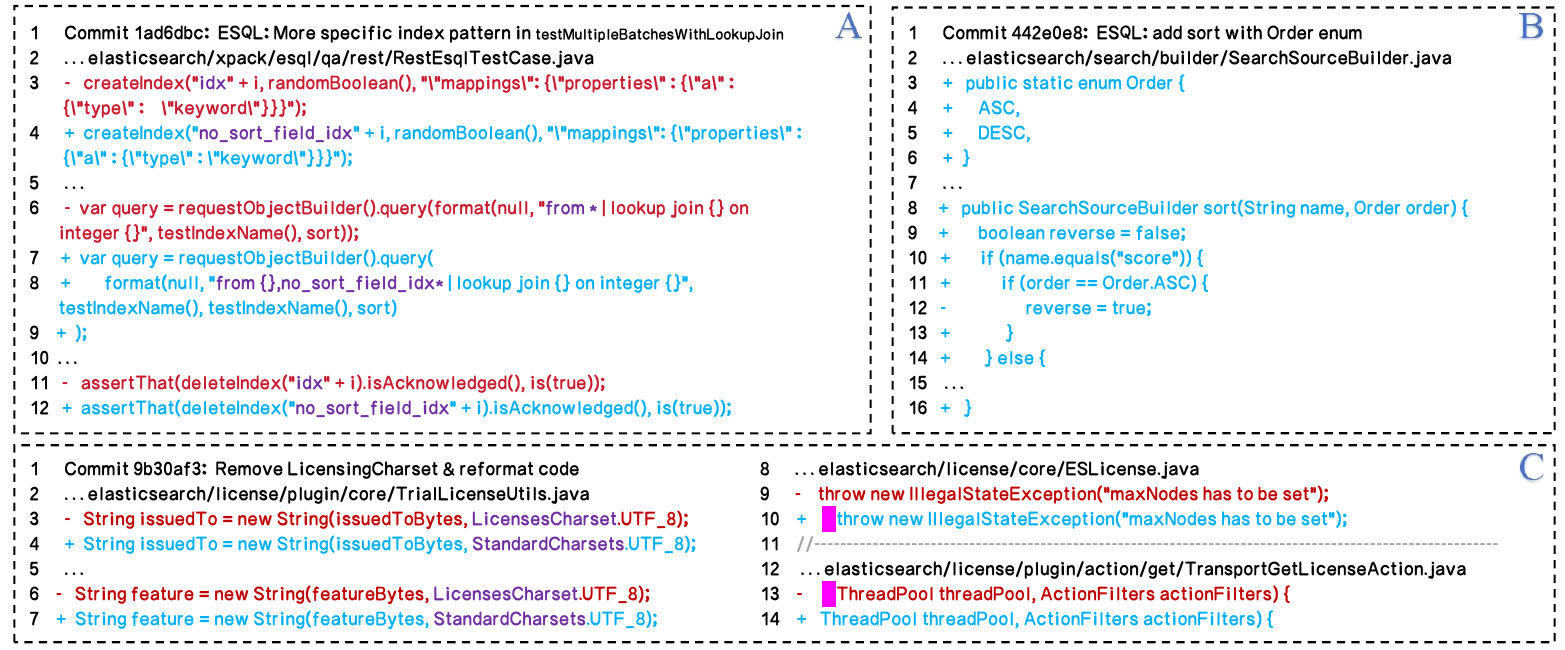}
    \caption{Motivating Examples}
    \label{fig:motivate_examples}
    \vspace{-0.3cm}
\end{figure*}

The examples in Figure~\ref{fig:motivate_examples}, demonstrate the inadequacy of existing studies and highlight the necessity for an advanced, semantically-aware approach.

Figure \ref{fig:motivate_examples}A (\textit{Example 1}) illustrates the changes made in commit \texttt{1ad6dbc} of the Elasticsearch project. 
This atom commit updates the index used in the \texttt{testMultipleBatchesWithLookupJoin} function from \texttt{"idx"} to \texttt{"no\_sort\_field\_idx"}. The commit includes three separate code changes: lines 3–4 create the new index, lines 6–9 query using the updated index, and lines 11–12 delete the index. These changes do not exhibit any obvious data or control dependencies, nor do they share structural similarities in their syntax. However, they are semantically related, as they collectively form a complete and independent development activity involving the creation, usage, and deletion of an index. This semantic coherence indicates that they belong to a single concern.

Prior graph-based approaches have attempted to capture such semantic relationships through learned graph embeddings over program dependency graphs, but remain fundamentally limited. In Example 1, the three added statements in the diff are: N1 (\texttt{createIndex("no\_sort\_field\_idx" + i, ...)}, in the first \texttt{for} loop), N2 (the updated \texttt{query} statement using the wildcard \texttt{"no\_sort\_field\_idx*"}, in the second \texttt{for} loop), and N3 (\texttt{\seqsplit{assertThat(deleteIndex("no\_sort\_field\_idx" + i)...)}}, in the third \texttt{for} loop). In the $\delta$-PDG, none of these three nodes share a data-flow edge: N2 and N3 use string literals independently rather than referencing any variable defined by N1. Each node resides in a different loop body, and there is no dependency path connecting them. UTango~\cite{li2022utango}'s GNN embeddings, which aggregate local structural neighborhoods, produce representations for N1, N2, and N3 that reflect only their own loop contexts; without connecting edges, clustering separates them. In HD-GNN~\cite{fan2024HD-GNN}'s entity reference graph, the absence of variable references between them prevents HD-GNN from establishing a statement-level link. Neither approach can infer that N1, N2, and N3 collectively implement a single development activity—extending the test to cover \texttt{no\_sort\_field\_idx} indices across their full lifecycle.

\begin{noleftmdframed}
\textbf{Challenge 1: Semantic Disconnect.} \textit{Precisely untangling commits requires inferring the underlying semantic purpose and grouping changes that lack explicit structural links but form a cohesive, complete development activity.}
\end{noleftmdframed}

\textcolor{black}{
To quantify the prevalence of semantic disconnect in real-world development, we conducted a preliminary empirical study on two popular C\# repositories from our evaluation datasets: Lean (13,130 commits) and Nancy (5,476 commits). We randomly sampled 374 and 360 commits respectively (95\% confidence level, 5\% margin of error). The first and second authors of this paper, with over five and ten years of software development experience, respectively, independently annotated each sampled commit with the assistance of dependency analysis tools. Disagreements between the two annotators were resolved through discussion and cross-validation until consensus was reached. The results show that 32.09\% and 36.11\% of commits are tangled commits, confirming their prevalence. More importantly, 70.86\% and 62.22\% of sampled commits contain code changes that are semantically related but lack explicit data or control dependencies—the semantic disconnect pattern illustrated in Example 1. This demonstrates that the need for deep semantic reasoning is a dominant characteristic of real-world commits, not merely an edge case, underscoring the necessity of our approach.
}

Figure \ref{fig:motivate_examples}B (\textit{Example 2}) shows another atom commit. Lines 3–6 define an \texttt{enum} class representing ordering options, and lines 8–16 implement a sorting method that utilizes this \texttt{enum} (only the \textit{if} block is shown in the figure, as the \textit{else} block has a similar structure). The code changes in lines 3–6 and lines 8–16 exhibit logical dependencies respectively, as they belong to the same class or method implementation. Furthermore, there is a clear data dependency: line 11 directly uses the value defined in the \texttt{enum} class introduced in lines 3–6. These changes are categorized as belonging to a single concern based on their \textbf{logical and data dependencies}, which are strong structural indicators for grouping. 

However, many concerns are defined by implicit signals, where such strong explicit links are absent. Figure \ref{fig:motivate_examples}C (\textit{Example 3.}) shows a tangled commit containing two distinct concerns: \textit{"remove LicensingCharset"} and \textit{"reformat code"}. The first concern (at lines 3 and 6) involves changes to the charset used in the \texttt{TrialLicenseUtils} class (only a subset of these changes is shown, as other charset modifications exhibit similar structure). Each of these changes is identical, replacing \texttt{LicensesCharset} with \texttt{StandardCharsets}. The affected statements are cloned code segments used to support different variables (\texttt{issueTo} and \texttt{feature}), resulting in \textbf{high textual and structural similarity} which is a crucial implicit signal for grouping. The variety of relationships between code changes—from the strong, explicit data flow in Example 2 to the subtle, implicit similarity here—presents an integration dilemma. 

\textcolor{black}{
Example 2 and the first concern of Example 3 illustrate that code changes can be related through either explicit dependencies (data/control flow, as in Example 2) or implicit dependencies (structural similarity, as in Example 3). Utango~\cite{li2022utango} addresses both via a sequential pipeline: GNN-based embeddings detect explicit structural dependencies first, and code clone detection handles implicit similarity for changes lacking explicit links. This pipeline correctly groups the changes in Example 2 (via the data-flow edge from the \texttt{enum} to the sorting method) and the cloned charset replacement statements in Example 3 (via structural similarity). However, the pipeline's fixed sequential order becomes a limitation when a commit contains changes that are semantically related but neither explicitly dependent nor structurally similar—as in Example 1, where three parts of code changes share a development purpose but reside in structurally dissimilar loop bodies with no dependency edges between them. In such cases, the first stage produces a poor clustering, and the clone detection step cannot recover it.
}

\begin{noleftmdframed}
\textbf{Challenge 2: Contextual Integration.} \textit{A unified approach must effectively integrate and weigh diverse forms of evidence—ranging from strong explicit dependencies (data/control flow) to subtle implicit relationships (semantics/similarity)—to accurately determine concern boundaries.}
\end{noleftmdframed}

The second concern in Figure \ref{fig:motivate_examples}C (\textit{Example 3}) focuses on reformatting code at lines 9, and 13, spanning two files. These changes represent typical \textbf{cosmetic edits}, which include syntactic formatting, refactoring, or non-functional textual modifications. Such edits are common in daily software development and usually belong to a single concern. However, existing approaches fail to detect and classify these changes correctly, as their rules~\cite{barnett2015Heuristic-rule, herzig2016impact}, features~\cite{dias2015EpiceaUntangler}, or graph representations~\cite{parțachi2020flexeme, li2022utango, fan2024HD-GNN} do not model cosmetic edits. Due to this oversight, they often fail to correctly isolate cosmetic edits, which can scatter across files and hide within larger functional concerns.
\textcolor{black}{
To be specific, the cosmetic changes in Example 3 are at lines 9 and 13 (spanning two files): pure whitespace/indentation reformatting edits with no functional impact. Prior approaches cannot handle such changes because cosmetic edits introduce no data flow, control flow, or entity references—the very signals that $\delta$-PDG-based methods (UTango~\cite{li2022utango}, Flexeme~\cite{parțachi2020flexeme}) and entity reference graph-based methods (HD-GNN~\cite{fan2024HD-GNN}) rely on. These changes are therefore invisible to the graph representations of prior approaches: they either do not appear as meaningful nodes in the graph, or appear as isolated nodes with no edges, and are arbitrarily merged with nearby changes or left ungrouped—without any recognition of their non-functional nature.
}

\begin{noleftmdframed}
\textbf{Challenge 3: Specialized Classification.} \textit{Existing methods lack the capability to accurately identify and isolate non-functional cosmetic changes, leading to incomplete untangling.}
\end{noleftmdframed}

Except for this, existing methods act as a black box, failing to provide the contextual reasoning behind their grouping decisions~\cite{parțachi2020flexeme, li2022utango, fan2024HD-GNN}. This lack of a clear, contextual rationale is a significant limitation in a task requiring complex contextual judgment. A robust solution should offer interpretability, explaining its reliance on specific evidence (e.g., "grouped due to semantic completion" or "separated due to clear data independence").

\begin{noleftmdframed}
\textbf{Challenge 4: Rationale and Explainability.} \textit{Existing untangling methods lack the capacity to provide a clear, contextual rationale for their grouping decisions.}
\end{noleftmdframed}

\subsection{Key Principles}
\label{key_ideas}
Based on the core challenges identified in Section \ref{challenges}, we propose $\textit{ColaUntangle}$ with an architecture founded on three key principles. These principles are designed to overcome the limitations by integrating deep semantic reasoning with a robust, multi-perspective decision-making process.

\subsubsection{\textit{Principle 1: LLM-driven Foundation Models}} 
To address Challenge 1 (Semantic Disconnect) and Challenge 3 (Specialized Classification), we decide to leverage the powerful capabilities of LLMs rather than traditional machine learning or neural network models to untangle commits.

\textcolor{black}{
Unlike graph-based methods such as UTango and HD-GNN which primarily capture structural proximity and syntactic patterns, LLMs possess pre-trained knowledge over massive code and natural language corpora, granting them a deep understanding of programming concepts, development activities, and coding conventions. This enables LLMs to perform high-level semantic reasoning—for instance, inferring changes that on the same entity collectively form a coherent development activity, even when these changes are structurally disconnected in the dependency graph (as illustrated in Example 1). Furthermore, LLMs can naturally recognize non-functional modifications such as formatting, import reordering, and documentation updates through their semantic understanding of code intent, accurately classifying them as cosmetic edits—a capability that heuristic rules, handcrafted features, and graph representations fundamentally lack (as illustrated in Example 3). 
}

\subsubsection{\textit{Principle 2: Explicit and Implicit Dependency}} Based on the motivating examples, we find that code changes can be grouped into different concerns due to various factors. To effectively guide the LLM's reasoning, we formalize the diverse observations from our motivating examples into a structured framework of evidence. We categorize the characteristics linking code changes into two distinct, orthogonal types: Explicit Dependencies and Implicit Dependencies.

\textit{Explicit dependencies} capture the direct relationships between code changes, such as data or control dependencies. In contrast, \textit{implicit dependencies} capture the underlying reasons why code changes without obvious structural or data relationships may still belong to the same concern, such as semantic similarity or shared development purpose. Specifically, we define the following rules:

\textbf{Explicit Dependencies Rules:}
\begin{itemize}
\item Code changes with data dependencies often belong to the same concern. 
\item Code changes with control dependencies often belong to the same concern. 
\end{itemize}

\textbf{Implicit Dependencies Rules:}
\begin{itemize}
\item Code changes with semantic similarity may belong to the same concern.
\item Code changes with high textual or structure similarity may belong to the same concern. 
\item Code changes introduced for cosmetic edits, such as syntactic formatting, refactoring, or non-functional textual modifications, often belong to the same concern.
\end{itemize}

Unlike prior studies that offered vague or indirect definitions of explicit and implicit dependencies, this formal set of rules provides a direct, comprehensive, and interpretable basis for untangling, enabling the LLM agents to reason directly over these untangling rules. 
The LLM agents can directly reason over these explicit principles. This design is fundamental to achieving both robust evidence integration (Challenge 2) and generating the clear, contextual rationale required for explainability (Challenge 4).
\textcolor{black}{Further, it is important to distinguish our concept of \textit{implicit dependencies} from \textit{hidden dependencies} as introduced by HD-GNN~\cite{fan2024HD-GNN}. \textit{Hidden dependencies} in HD-GNN are latent variables learned by GNNs through hierarchical embedding aggregation over entity reference graphs. While these learned representations can capture some non-obvious patterns for classifying code change pairs, they are abstract numerical features without clear semantic meaning—one cannot explain what specific type of relationship a hidden dependency represents or why it leads to a particular grouping decision. In contrast, our \textit{implicit dependencies} are explicitly formalized as interpretable rules that capture clearly defined semantic-level relationships. Each rule provides a concrete, comprehensible basis for reasoning, enabling LLM agents to not only make accurate untangling decisions but also generate transparent explanations for their groupings. This distinction is central to our contribution: by transforming the vague notion of non-obvious relationships into a principled and interpretable evidence framework, we enable systematic reasoning over the full spectrum of relationships governing commit untangling.}

\subsubsection{\textit{Principle 3: Collaborative Consultation}} 
\textcolor{black}{Recognizing that untangling requires resolving conflicts between the diverse evidence defined in Principle 2, and to address Challenge 2 (Contextual Integration) and Challenge 4 (Rationale and Explainability), we implement a multi-agent collaborative consultation process comprising distinct roles and iterative negotiation. Specifically, we define two types of worker agents: one specializing in analyzing explicit dependencies between code changes and the other focusing on implicit dependencies. Additionally, we introduce a reviewer agent that synthesizes and integrates the outputs from both worker agents. These three agents engage in multiple rounds of collaborative consultation, producing the final untangling result upon reaching consensus. Unlike the sequential, rigid pipelines of prior methods (e.g., UTango's embedding-then-clone-detection approach), our multi-agent architecture enables concurrent analysis of both explicit and implicit dependencies, with iterative reconciliation through the reviewer agent. This design allows dynamic, context-aware balancing of diverse evidence types. Moreover, all agents are required to provide natural language explanations for their decisions, enabling transparent and human-verifiable untangling results—a capability absent in black-box GNN-based approaches.}

\section{Methodology}

\label{sec:Approach}
\begin{figure*}
    \centering
    \includegraphics[width=0.98\linewidth]{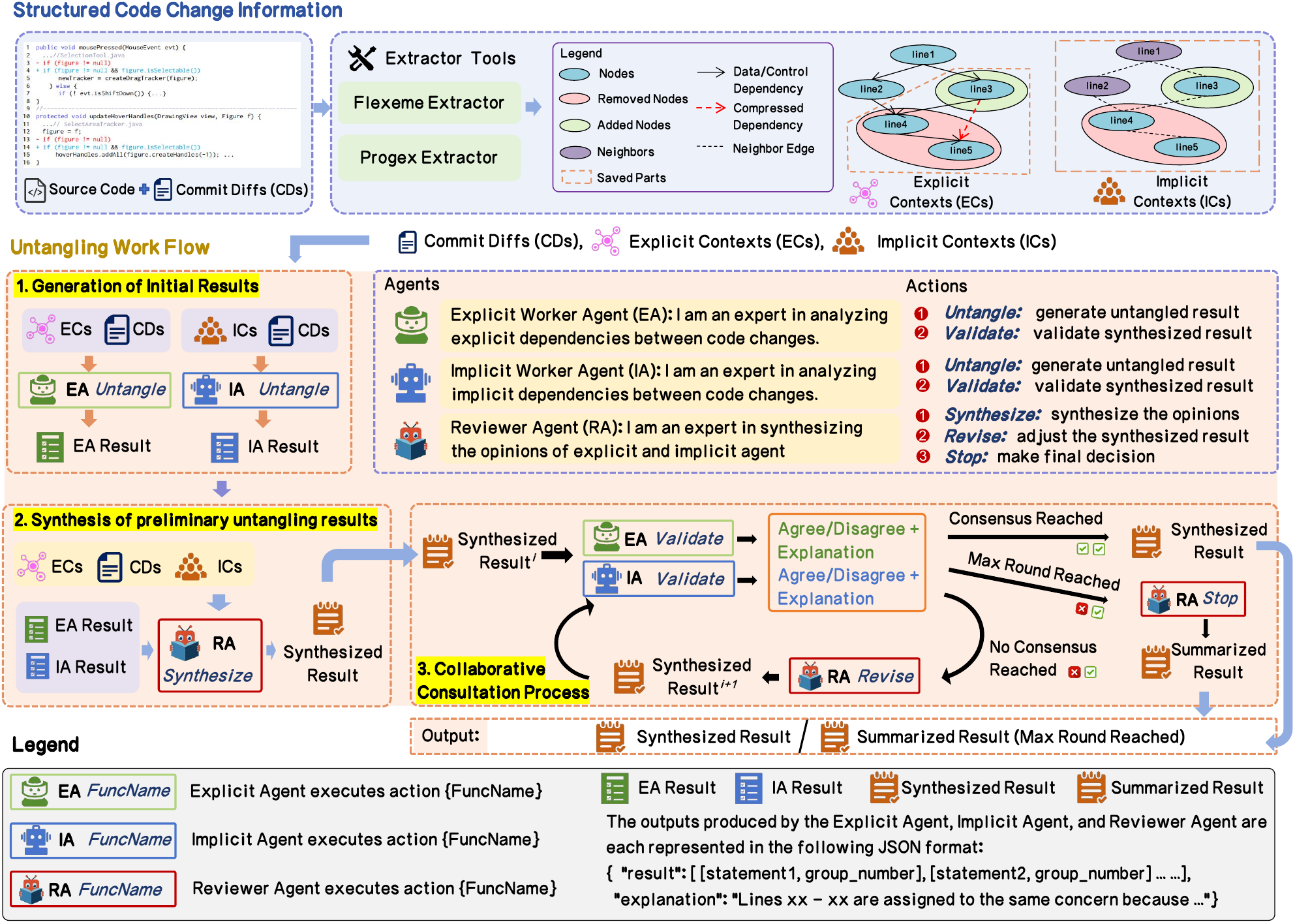}
   
    \caption{\textcolor{black}{Overall Workflow of \textit{ColaUntangle}. The input is a tangled commit consisting of source code files. The output is a concrete untangling partition, where each code change is assigned a concern label. During the workflow, both the \textit{synthesized result} (produced by RA after integrating EA and IA's results) and the \textit{summarized result} (produced by RA after collaborative consultation reaches consensus or the maximum number of rounds) are concrete untangling partitions where each code change (statement) is assigned a concern label (e.g., Concern 1, Concern 2, ...) with explanation.}}
    \label{fig:approach_overview}
\end{figure*}

We propose \textit{ColaUntangle}, a collaborative framework that leverages LLM-driven agents to untangle commits by identifying both explicit and implicit dependencies among code changes. As shown in Figure~\ref{fig:approach_overview}, \textit{ColaUntangle} first extracts structured code change information, including commit diffs and contexts derived from source code. It then employs a multi-agent consultation process, where agents iteratively analyze dependencies and synthesize untangling decisions until reaching consensus.

\subsection{Structured Code Change Information}
To enrich the contextual information available for dependency analysis, we first extract \textit{explicit contexts} and \textit{implicit contexts} from the program dependency graphs (PDGs) from the source code. Our method centers on building a consolidated graph structure sensitive to code changes, followed by targeted reduction and compression.

A \textit{program dependency graph (PDG)} \cite{ferrante1987PDG} is a directed graph that consists of a set of nodes \textit{N} and a set of edges \textit{E}, where each node n $\in$ \textit{N} represents a program statement or a conditional expression; and each edge e $\in$ \textit{E} represents the data flow or control flow among the statements within a single version of the code~\cite{ferrante1987PDG, le2014patch}. However, PDGs can only describe the dependency relationships between statements of an individual program.

To simultaneously reflect statement dependencies before and after program modification, we first considered the capabilities of the Multi-version Program Dependency Graph ($\delta$-PDG) \cite{parțachi2020flexeme, kim2006program}, which is conceptually defined as the disjoint union of all nodes and edges from the PDGs of versions $i$ and $j$. We utilized this structural concept to construct our Commit-Level Merged Graph ($G_{Merged}$). However, $G_{Merged}$ is often massive, presenting a critical challenge for context-aware analysis with LLMs due to input length constraints, leading to inefficient token utilization and potentially diffusing the model's focus away from the core changes. Therefore, we further explore the specialized extraction of contexts from $G_{Merged}$ to derive representations optimized for LLM consumption, namely the Explicit Context ($G_{Explicit}$) and the Implicit Context ($G_{Implicit}$).

We first construct the \textbf{Commit-Level Merged Graph ($G_{Merged}$)}, which serves as the consolidated foundation for our context extraction. As illustrated in Algorithm \ref{alg:ex_im_contexts_generation}, the construction process begins by taking the PDGs before ($\mathcal{G}_B$) and after ($\mathcal{G}_A$) the code change for every affected file, along with the commit's line-level difference ($\mathcal{D}$). \textcolor{black}{In practice, we construct the PDGs using the Flexeme extractor\footnote{\url{https://github.com/PPPI/Flexeme}} for C\# projects and the Progex extractor\footnote{\url{https://github.com/ghaffarian/progex}} for Java projects.} Using $\mathcal{D}$, we mark nodes in $\mathcal{G}_B$ corresponding to deleted lines as Deleted, and nodes in $\mathcal{G}_A$ corresponding to added lines as Added.

Following this initial marking, all file-level marked PDGs are aligned and merged into the single $G_{Merged}$ structure, where common unchanged nodes are consolidated using a fuzzy matching strategy (Lines 1-8). This construction relies on a robust fuzzy node matching strategy to identify semantically equivalent nodes between the two versions. Specifically, a node from $\mathcal{G}_B$ is matched with one from $\mathcal{G}_A$ if their node type, their structural context similarity, and their label content similarity all exceed defined fuzziness cutoffs ($\tau$), which is set to 1.0 following the practice of Flexeme~\cite{parțachi2020flexeme}. Nodes successfully matched are consolidated as Unchanged Nodes. The resulting $G_{Merged}$ is constructed by consolidating the matched nodes (Unchanged Nodes) and their edges, retaining the unmatched nodes from $\mathcal{G}_B$ (Deleted nodes) and $\mathcal{G}_A$ (Added nodes), along with all their respective edges. This approach yields $G_{Merged}$, which represents the disjoint union of nodes and edges across both versions, with common elements consolidated, thereby effectively capturing the complete dependency scope of the committed changes.

\begin{algorithm}[H]
\footnotesize
\caption{Explicit and Implicit Contexts Generation}
\label{alg:ex_im_contexts_generation}
\begin{algorithmic}[1]
\Require \parbox{\dimexpr\linewidth-1.5em}{\raggedright $\mathcal{D}$: Commit diff list (lines added/deleted); $\mathcal{G}_{B}, \mathcal{G}_{A}$: PDGs of modified files (Before, After); $\tau$: Node matching fuzziness cutoffs.}
\Ensure \parbox{\dimexpr\linewidth-1.5em}{\raggedright $G_{Explicit}$: Explicit Context.}
\Ensure \parbox{\dimexpr\linewidth-1.5em}{\raggedright $G_{Implicit}$: Implicit Context.}

\State $\mathcal{L}_{files} \leftarrow []$ \Comment{Initialize list for file-level Merged PDGs}

\For{each modified file $F$}
    \State $\mathcal{D}_F \leftarrow \text{DIFF\_FOR\_FILE}(\mathcal{D}, F)$
    \State $G_{B,F}^{\prime} \leftarrow \text{MARK\_PDG}(\mathcal{G}_{B}[F], \mathcal{D}_F)$ \Comment{Mark deleted nodes according to commit diffs}
    \State $G_{A,F}^{\prime} \leftarrow \text{MARK\_PDG}(\mathcal{G}_{A}[F], \mathcal{D}_F)$ \Comment{Mark added nodes according to commit diffs}
    \State $G_{\Delta,F} \leftarrow \text{MERGE\_MARKED}(G_{B,F}^{\prime}, G_{A,F}^{\prime}, \tau)$ \Comment{Generate Merged PDG for each file}
    \State $\mathcal{L}_{files} \leftarrow \mathcal{L}_{files} \cup \{ G_{\Delta,F} \}$
\EndFor

\State $G_{Merged} \leftarrow \text{MERGE\_FOR\_COMMIT}(\mathcal{L}_{files})$ \Comment{Combine all files into a single graph}
\State $N_{Changed} \leftarrow \{ N \in G_{Merged} \mid N.\text{color} \in \{\text{`red'}, \text{`green'}\} \}$

\State $N_{Implicit} \leftarrow N_{Changed} \cup \{ \text{neighbors}(N) \mid N \in N_{Changed} \}$  \Comment{Implicit Context Generation}
\State $G_{Implicit} \leftarrow G_{Merged}.\text{subgraph}(N_{Implicit})$

\State $G_{Explicit} \leftarrow G_{Merged}.\text{subgraph}(N_{Changed})$
\For{each pair $(S, T)$ in $N_{Changed} \times N_{Changed}$} \Comment{Explicit Context Generation}
    \If{$S \neq T$ \textbf{and} $\text{PATH\_EXISTS}(G_{Merged}, S, T)$}
        \State $\text{ADD\_EDGE}(G_{Explicit}, S, T, \text{type=`shortcut'})$
    \EndIf
\EndFor

\State \Return $G_{Explicit}, G_{Implicit}$
\end{algorithmic}
\end{algorithm}

For most modification operations, the changed parts account for only a small proportion of the entire program. To reduce unnecessary information, we introduce the concept of \textit{explicit contexts}. The Explicit Context (i.e., $G_{Explicit}$) is a highly compressed graph designed to capture the logical dependency flow solely among the modified statements, disregarding intervening unchanged code. To construct $G_{Explicit}$, we first initialize a subgraph containing only the changed nodes ($N_{\text{Changed}}$), as shown in Algorithm \ref{alg:ex_im_contexts_generation} (Line 10). We then apply dependency compression by adding shortcut edges: for every distinct pair of nodes $(S, T) \in N_{\text{Changed}}$ that have a dependency path in the complete $G_{Merged}$, a direct shortcut edge is added between $S$ and $T$ in $G_{Explicit}$ (Lines 14-18). Although $G_{Explicit}$ loses all surrounding context, it effectively preserves the critical data and control dependencies between the modified statements, offering a minimal, dependency-centric view of the change impact.

Conversely, the Implicit Context (i.e., $G_{Implicit}$ ) is designed to retain neighbor code information without applying compression. $G_{Implicit}$ is generated by extracting the subgraph that includes all changed nodes ($N_{\text{Changed}}$) and their direct one-hop neighbors in $G_{Merged}$ (Lines 11-12). $G_{Implicit}$ strictly preserves only the original PDG edges present among these nodes and contains no shortcut edges. By including the immediate neighbors, $G_{Implicit}$ offers a larger, more comprehensive view of the surrounding code and structural environment than $G_{Explicit}$, enabling LLMs to leverage richer semantic information for accurate analysis.

\begin{figure}
    \centering
    \setlength{\abovecaptionskip}{0.1cm}
    \includegraphics[width=0.9\linewidth]{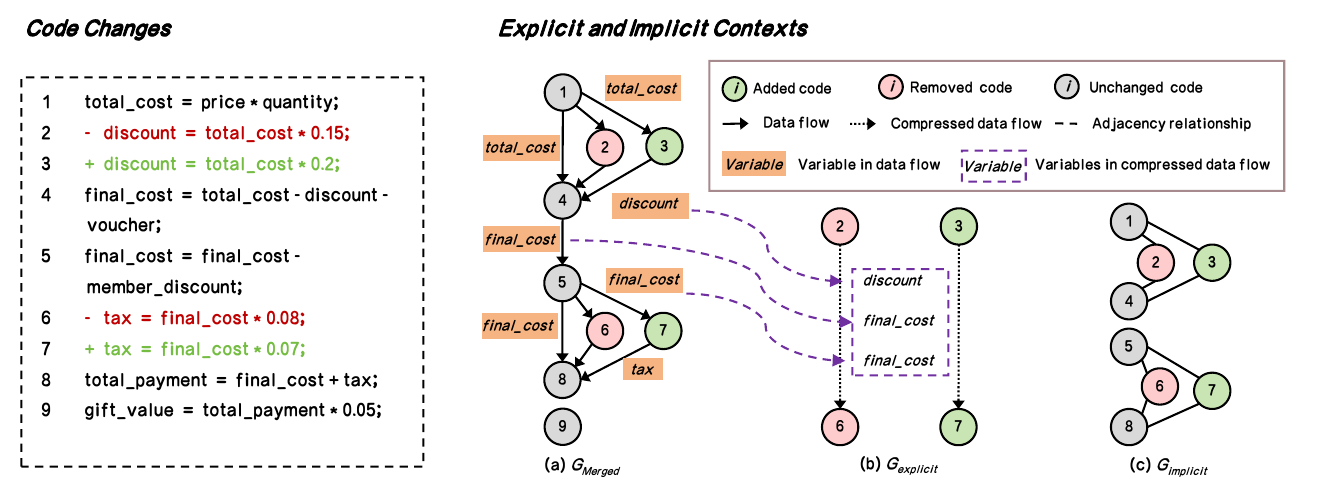}
    \caption{Code Change Example with Explicit and Implicit Contexts}
    \label{fig:structured_code_change_information_example}
    \vspace{-0.3cm}
\end{figure}

Figure \ref{fig:structured_code_change_information_example} illustrates a code change scenario involving two statement replacements (at lines 2 and 6). Figure \ref{fig:structured_code_change_information_example} (a) shows the $G_{Merged}$ of this code snippet, where red nodes represent the statements before the change, green nodes represent the modified statements, and gray nodes denote the unchanged statements. Figure \ref{fig:structured_code_change_information_example} (b) presents the \textit{explicit contexts ($G_{explicit}$)}, which are obtained by compressing the $G_{Merged}$ in Figure \ref{fig:structured_code_change_information_example} (a). Specifically, we retain the changed nodes and their direct dependency edges; additionally, if two changed nodes lack a direct dependency but are connected via a dependency path (channel), we add a compressed edge between them. For example, nodes 2 and 6 have no direct dependency but are connected via a path through nodes 4 and 5, resulting in a compressed edge between them. This compressed edge is labeled with the variables involved in the data flow along the original dependency path. Figure \ref{fig:structured_code_change_information_example} (c) shows the \textit{implicit contexts ($G_{implicit}$)}, which retain both the changed nodes and their immediate neighbors (within one-hop dependencies). In this example, node 9 is excluded because it does not have a direct dependency with any of the changed nodes.

\subsection{Multi-Agent Collaborative Consultation}
As shown in Figure~\ref{fig:approach_overview}, the multi-agent collaborative consultation involves three types of agents: two worker agents specializing in mining explicit and implicit dependencies between code changes, respectively, and a reviewer agent responsible for reviewing and synthesizing the untangling results. The process begins with the two worker agents generating an initial untangling of a tangled commit based on its structured code information. The reviewer agent then synthesizes these results, initiating the collaborative consultation. In each iteration, the worker agents provide feedback on the current synthesized result, which the reviewer agent uses to revise synthesized result accordingly. This iterative process continues until the agents reach a consensus or a maximum number of rounds is reached.

\subsubsection{Agent Building}
\begin{figure}[htbp]
    \centering
    \setlength{\abovecaptionskip}{0.1cm}
    \begin{subfigure}[t]{0.49\linewidth}
        \centering
        \includegraphics[width=\linewidth]{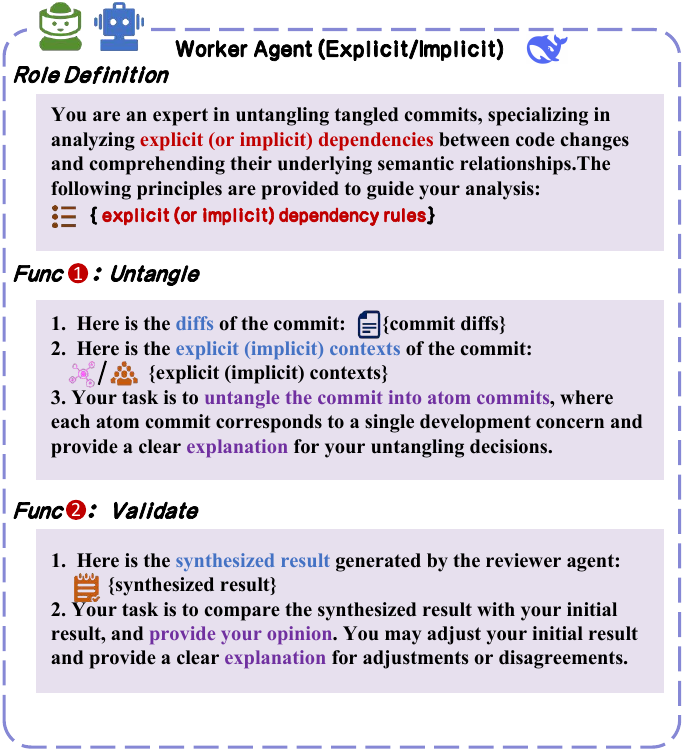}
        \caption{Worker Agent}
        \label{fig:agent_build_worker}
    \end{subfigure}
    \hfill 
    \begin{subfigure}[t]{0.49\linewidth}
        \centering
        \includegraphics[width=\linewidth]{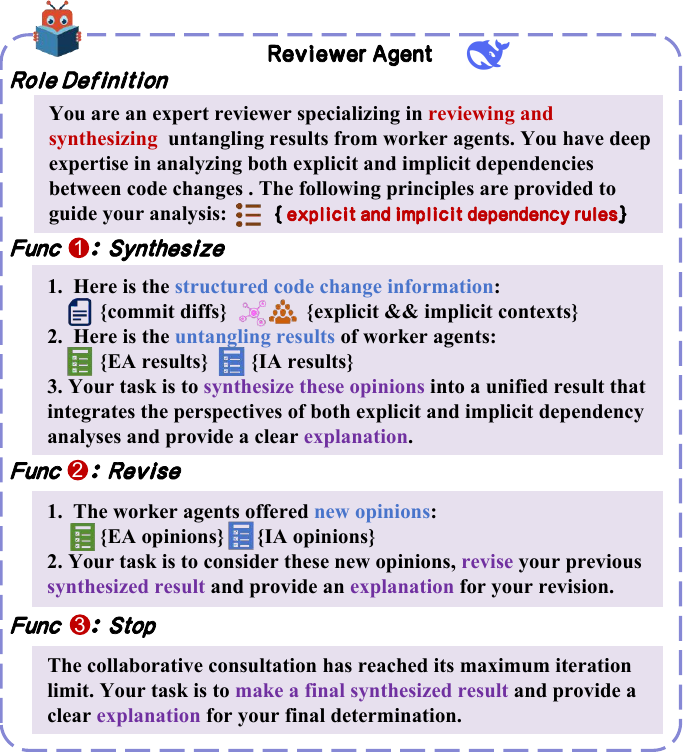}
        \caption{Reviewer Agent}
        \label{fig:agent_build_reviewer}
    \end{subfigure}
    
    \caption{Agents in Multi-Agent Collaboration}
    \label{fig:multi_agent_build_process}
    \vspace{-0.3cm}
\end{figure}
\textit{ColaUntangle} builds three types of agents. The first two are worker agents, each specializing in untangling commits based on either explicit or implicit dependencies between code changes, as illustrated in Figure~\ref{fig:agent_build_worker}. The third is a reviewer agent, which specializes in reviewing and synthesizing the results from the worker agents, as shown in Figure~\ref{fig:agent_build_reviewer}.

\textbf{Worker Agent.} Based on the explicit and implicit dependency rules summarized in Section~\ref{key_ideas}, we define two types of worker agents, as illustrated in Figure~\ref{fig:agent_build_worker}. The worker agents specialize in untangling commits by analyzing explicit and implicit dependencies between code changes, respectively. We refer to the explicit worker agent as \textit{EA} and the implicit worker agent as \textit{IA}.
We define the roles of these agents using the following role definition prompt: ``\textit{You are an expert in untangling tangled commits, specializing in analyzing explicit (or implicit) dependencies between code changes and comprehending their underlying semantic relationships}''. We also provide each agent with the corresponding explicit or implicit dependency rules (concluded in Section \ref{key_ideas}) to guide their analysis.
Each worker agent has two main functions. The first function is ``\textit{untangle}''. EA untangles tangled commits based on the inputs of \textit{commit diffs} ($\mathcal{D}$) and \textit{explicit contexts} ($G_{Explicit}$), while IA performs untangling based on \textit{commit diffs} ($\mathcal{D}$) and \textit{implicit contexts} ($G_{Implicit}$). Both agents are required to generate explanations for their decisions to facilitate understanding of their underlying criteria.
The second function is ``\textit{validate}''. In this function, each worker agent compares the synthesized result generated by the reviewer agent (described in the next section) with its own untangling result and provides its opinion on the synthesized result. This includes indicating agreement or disagreement and, if disagreeing, providing a revised untangling result along with an explanation.

\textbf{Reviewer Agent}. The untangling results produced by the two worker agents may conflict due to their different expertise and the presence of diverse explicit and implicit dependencies within code changes. To resolve these conflicts, we define a reviewer agent, referred to as \textit{RA}, which specializes in reviewing and synthesizing the untangling results from the worker agents, as illustrated in Figure~\ref{fig:agent_build_reviewer}.
We define the reviewer agent’s role using the following role definition prompt: ``\textit{You are an expert reviewer specializing in reviewing and synthesizing untangling results from worker agents. You have deep expertise in analyzing both explicit and implicit dependencies between code changes}''. To support its role, we provide RA with both explicit and implicit dependency rules to guide its synthesis process.
RA has three main functions: \textit{synthesize}, \textit{revise} and \textit{stop}. In the \textit{synthesize} function, RA synthesizes the results from EA and IA based on the structured code change information ($\mathcal{D}$, $G_{Explicit}$, and $G_{Implicit}$). In the \textit{revise} function, RA revises its synthesized result based on the new opinions generated by the two worker agents during validation. Finally, in the \textit{stop} function, when the collaborative consultation process reaches its maximum iteration limit, RA produces the final untangling result. For each function, RA is required to provide clear explanations for its decisions.

\subsubsection{Untangling Workflow} 
The workflow of \textit{ColaUntangle} is illustrated in Figure~\ref{fig:approach_overview}. The collaboration begins with the worker agents (EA and IA) providing their initial untangling results. Subsequently, the reviewer agent (RA) synthesizes these results into a unified untangling outcome. This initiates an iterative collaborative consultation process, during which EA and IA express their opinions on the current synthesized result. RA then considers their feedback and revises the synthesized result accordingly. This iterative process continues until EA and IA reach a consensus on RA's synthesized result or a predefined maximum number of rounds is reached. Specifically, the untangling workflow comprises three phases: \textbf{generation of initial reviews}, \textbf{synthesis of preliminary untangling results}, and \textbf{collaborative consultation process}.

In the first phase, \textbf{generation of initial reviews}, EA and IA are instructed to generate their preliminary untangling results $r_{EA}$ and $r_{IA}$ using their \textit{Untangle} function, based on the commit diffs ($\mathcal{D}$) and the explicit ($G_{Explicit}$) or implicit ($G_{Implicit}$) contexts:
\[r_{EA}=EA.Untangle(Prompt(\mathcal{D},G_{Explicit}))\]

\[r_{IA}=IA.Untangle(Prompt(\mathcal{D},G_{Implicit}))\]

\textcolor{black}{In the second phase, \textbf{synthesis of preliminary untangling results}, RA uses its \textit{Synthesize} function to review and integrate the results from EA and IA into a synthesized result $r_{RA}$. The synthesized result is a concrete untangling partition that assigns each code change to a specific concern group (e.g., Concern 1, Concern 2, ...), produced by RA after comprehensively considering both the explicit and implicit dependency analyses from EA and IA.}

\[r_{RA}=RA.Synthesize(Prompt(r_{EA},r_{IA},\mathcal{D},G_{Explicit},G_{Implicit}))\]

\begin{algorithm}[htbp]
\footnotesize
\caption{Collaborative Consultation}
\label{alg:collaborative_consultation}
\begin{algorithmic}[1]
\Require Experts $EA$, $IA$, $RA$; initial synthesized result $r_{RA}$; maximum attempts $t$; prompt template $Prompt$
\Ensure Final result $r_f$
\State $consensus \gets False$, $round \gets 0$ \Comment{Initialize variables}
 
\While{$consensus = False$ \textbf{and} $round < t$} \Comment{Iterative consultation}
    \State $round \gets round + 1$
    \State $consensus \gets True$
    
    \State $EA\_opinion \gets EA.validate(Prompt(r_{RA}))$
    \State $IA\_opinion \gets IA.validate(Prompt(r_{RA}))$
    \If{$EA\_opinion.agree = False$ \textbf{or} $IA\_opinion.agree = False$}
        \State $consensus \gets False$
        \If{$round = t$}
           
            \State $r_{RA} \gets RA.stop(EA\_opinion, IA\_opinion)$
        \Else
           
            \State $r_{RA} \gets RA.revise(EA\_opinion, IA\_opinion)$
        \EndIf
    \EndIf
\EndWhile
\State $r_f \gets r_{RA}$
\State \Return $r_f$
\end{algorithmic}
\end{algorithm}

The final phase, \textbf{collaborative consultation process}, is iterative as shown in Algorithm~\ref{alg:collaborative_consultation}. During each round $j$, the worker agents EA and IA evaluate the current synthesized result and express their agreement or disagreement. In cases of disagreement, they are required to provide detailed rationales. To be specific, in round $j$-th:
\[EA\_opinion^j = EA.validate(Prompt(r_{RA}^{j-1}))\]
\[IA\_opinion^j = IA.validate(Prompt(r_{RA}^{j-1}))\]
If both agents reach unanimous agreement, the iterative process terminates (next loop condition is not satisfied). Otherwise, if the maximum round limit has not been reached, RA uses its \textit{Revise} function to consider the opinions from EA and IA, refining the current synthesized result to produce a new synthesized result accordingly:
\[r_{RA}^j=RA.revise(Prompt(EA\_opinion^{j},IA\_opinion^{j}))\]

\textcolor{black}{
If the maximum round limit is reached without achieving consensus, RA uses its \textit{Stop} function to produce a \textit{summarized result}—a final, concrete untangling partition that comprehensively considers all accumulated opinions from EA and IA across the consultation rounds. Note that both the \textit{synthesized result} (produced during iterative consultation) and the \textit{summarized result} (produced at termination) share the same form: a concrete untangling partition where each code change is assigned to a specific concern group, along with explanations for the grouping decisions.
}
\[r_{RA}^j=RA.stop(Prompt(EA\_opinion^{j},IA\_opinion^{j}))\]
In the next round, the worker agents once again express their opinions towards the new synthesized result and try to convince the RA. The RA considers their statements and further revises the result until all worker agents reach a consensus or the maximum number of rounds is reached.

\section{Experimental Setups}

\subsection{Datasets}  \label{subsec:dataset}{}
 We use two datasets that have been widely used in previous work: 1) A C\# dataset used in \cite{chen2022attributedgraph,herzig2016impact,parțachi2020flexeme,li2022utango,fan2024HD-GNN}, which contains 1,612 tangled commits (each of which has $\le$ 3 concerns) across 9 GitHub projects. 2) A Java dataset used in \cite{li2022utango,fan2024HD-GNN}. The Java dataset contains over 14k tangled commits in 10 GitHub repositories, where the number of concerns in a commit ranges from 2-32. Both datasets were constructed by first selecting atomic commits based on criteria such as temporal proximity, namespace similarity, and file coupling, and then artificially tangling them via git cherry-picking to simulate realistic tangled commit scenarios.

\subsection{Baselines}

\textbf{Commit Untangling-specific Baselines}.
The Baselines used for C\# dataset are: 

\begin{itemize}[leftmargin=0pt,itemindent=0pt,labelwidth=0pt,labelsep=0pt,noitemsep]
\item[] \textit{Barnett et al.} \cite{barnett2015Heuristic-rule}: A heuristic rule-based approach considering def-use, use-use, and same-enclosing-method relations among code changes. 

\item[] \textit{Herzig et al.} \cite{herzig2016impact}: A heuristic rule-based approach combining various Confidence Voters and build a triangle partition matrix to untangle the commits. 

\item[] \textit{Flexeme} \cite{parțachi2020flexeme}: A graph clustering-based approach builds a $\delta$-NFG (proposed in \textit{Flexeme}) of commits and then applies agglomerative clustering on this $\delta$-NFG to untangle the commits. 

\item[] \textit{$\delta$-PDG+CV} \cite{parțachi2020flexeme}: A variant of \textit{Flexeme} by applying \textit{Herzig et al.}'s confidence voters directly to $\delta$-PDG (proposed in \textit{Flexeme}). 

\item[] \textit{UTango} \cite{li2022utango}: A graph clustering-based approach builds a $\delta$-PDG from commits and then applies GNN and agglomerative clustering on this $\delta$-PDG to untangle the commits. 

\item[] \textit{HD-GNN} \cite{fan2024HD-GNN}: A graph clustering-based approach builds an entity reference graph from commits and then applies GNN to aggregate the embeddings of nodes and edges in the graph and classify the relation between the code changes. 
\end{itemize}

The baselines used for Java dataset are: \textit{Barnett et al.}~\cite{barnett2015Heuristic-rule}, \textit{Herzig et al.}~\cite{herzig2016impact}, \textit{UTango}~\cite{li2022utango},  and \textit{HD-GNN}~\cite{fan2024HD-GNN} mentioned above, as all these approaches support analysis of Java program files. \textit{SmartCommit}~\cite{shen2021smartcommit} is a graph clustering-based interactive method that builds a Diff Hunk Graph and applies a graph partitioning algorithm to untangle commits. We also include \textit{SmartCommit} and its baselines, \textit{Base-1} (a rule-based approach that groups all changes into a single concern) and \textit{Base-2} (which groups changes by file), in our comparison.

\textbf{LLM-based Baselines}. We incorporate baselines utilizing single LLM for both C\# and Java datasets. We employ two different prompt strategies: zero-shot and zero-shot CoT. Zero-shot refers to prompting LLM to perform a task without providing examples, zero-shot CoT directly incorporates the prompt \textit{Let's think step by step} after a question to facilitate inference \cite{kojima2022zeroshot}.

\subsection{Metrics}

We use two evaluation metrics: $Accuracy^{c}$ proposed by Utango \cite{li2022utango} and $Accuracy^{a}$ proposed by Flexeme \cite{parțachi2020flexeme}. 
$Accuracy^{c}$ is defined as the percentage of the changed statements that are labeled with a correct cluster/concern in all the statements in a commit: \[Accuracy^{c}=\frac{\#changed\ stmts\ w.\ correct\ clusters}{\#all\ changed\ stmts\ in\ commit} \] 

$Accuracy^{a}$ is similar to $Accuracy^{c}$ except that it considers all statements: \[Accuracy^{a}=\frac{\#stmts\ w.\ correct\ clusters}{\#all\ stmts\ in\ commit} \]

\subsection{Implementation Details}

We construct the PDGs of C\# and Java dataset with the assistance of the Flexeme extractor\footnote{https://github.com/PPPI/Flexeme} and Progex extractor\footnote{https://github.com/ghaffarian/progex} respectively. We utilize the widely adopted and publicly available DeepSeek-V3 model \cite{liu2024deepseek} from the Azure OpenAI service\footnote{https://learn.microsoft.com/en-us/azure/ai-services/openai/} as the reasoning engine in our experiments, which is also used for the LLM-based baselines. All experiments are conducted under a zero-shot setting. All LLMs used in our experiments are accessed via their online APIs, all decoding parameters are kept at their default values. From an initial evaluation on 50 randomly selected tangled commits, we observed an average of 1.27 consultation rounds to reach consensus. Thus, we set the maximum number of rounds $t$ to 3, aligning with ColaCare \cite{wang2025colacare} (a medical multi-agent collaborative consultation model), and balancing iterative refinement with computational efficiency. In terms of computational cost, our approach incurs approximately \$2.92 per 100 examples (around \textcent2.92 per example), with an average untangling time of 71.63 seconds per example (Context Building: 9.79 s; Generation of Initial Results: 12.23 s; Synthesis of Preliminary Untangling Results: 14.68 s; Collaborative Consultation Process: 34.93 s)\footnote{We acknowledge that our approach requires longer inference time compared to prior methods. However, it eliminates the need for manual heuristic rule design, feature engineering, and model training, making it relatively cost-effective given its significant improvement in untangling accuracy.}. We conduct experiments on a Linux server running Ubuntu 22.04 system, equipped with an Intel® Core™ i7 processor 14700k, 32GB RAM.

\section{Experimental Results}

To evaluate \textit{ColaUntangle}, we seek to answer the following research questions:

\begin{itemize}[leftmargin=*,labelsep=0pt,noitemsep]
\item \textbf{[RQ1. Overall Performance]} How well does \textit{ColaUntangle} perform in comparison with the state-of-the-art commit untangling approaches?
\item \textbf{[RQ2. Ablation Study]} What is the contribution of each proposed module to the performance?

\textcolor{black}{\item \textbf{[RQ3. Sensitivity Analysis]} How sensitive is \textit{ColaUntangle}'s performance to different LLMs and key design parameters?}
\end{itemize}

\begin{table*}[h!t]
\setlength{\abovecaptionskip}{0.1cm}
\caption{Performance on the C\# Dataset ($Accuracy^c$\%/$Accuracy^a$\%)}
\small
\label{tab:c_overall_perfomance}
\begin{threeparttable}
\begin{tabular}{lcccccccccc}
\toprule
Method               & CL             & CM             & HF             & HU             & LE             & NA             & NJ             & NI             & RS             & Overall             \\
\hline
\multicolumn{11}{c}{\cellcolor[HTML]{EFEFEF}\textit{Previous Untangling Approaches}}                                                                                                           \\
\hline
Barnett et al.       & 14/20          & */20           & 12/15          & 13/25          & 7/16           & */9            & 7/13           & 10/14          & 10/13          & 7/13           \\
Herzig et al.        & 28/58          & */65           & 28/62          & 27/53          & 27/66          & */63           & 28/64          & 26/57          & 31/70          & 28/66          \\
\textit{$\delta$-PDG + CV} & 34/81          & */90           & 36/86          & 30/69          & 35/78          & */83           & 34/78          & 37/94          & 33/64          & 35/84          \\
Flexeme              & 34/87          & */70           & 33/77          & 33/70          & 35/80          & */87           & 27/62          & 32/80          & 34/86          & 34/83          \\
UTango               & 46/90          & \textit{*/*}   & 46/88          & 44/89          & 46/89          & \textit{*/*}   & 41/83          & 46/91          & 46/88          & 46/90          \\
HD-GNN               & 52/91          & \textit{*/*}   & 52/90          & 50/90          & 50/91          & \textit{*/*}   & 45/86          & 54/93          & 47/90          & 50/90          \\
\hline
\multicolumn{11}{c}{\cellcolor[HTML]{EFEFEF}\textit{\textit{ColaUntangle} Variants}}                                                                                                                       \\
\hline
$LLM_{ZeroShot}$         & 62/90          &  68/90              & 64/93          & 44/66          & 58/92          & 52/88          & 68/97          & 63/91          & 60/92          & 60/90          \\
$LLM_{ZeroShotCoT}$     & 63/90          &  68/90               & 64/93          & 45/66          & 59/92          & 54/88          & 68/97          & 65/92          & 60/92          & 61/90          \\
\textit{ColaUntangle}$_{no\_comments}$         & 74/94          & \textbf{74/92}          & 80/96          & 65/76          & 70/94          & 58/90          & 66/97          & 68/93          & 60/93          & 70/93          \\
\textit{ColaUntangle}                 & \textbf{75/94}   & 72/90    & \textbf{80/96} & \textbf{65/77} & \textbf{70/94} & \textbf{65/91} & \textbf{75/98} & \textbf{74/93} & \textbf{64/93} & \textbf{72/93} \\ 
\bottomrule
\end{tabular}
\begin{tablenotes}
\footnotesize
\item * refers no available data point.\\
\textit{ColaUntangle}$_{no\_comments}$ refers to running our method on the dataset excluding comments.
\end{tablenotes}
\end{threeparttable}
\vspace{-0.3cm}
\end{table*}

\begin{table}[h!t]
\setlength{\abovecaptionskip}{0.1cm}
\caption{Performance on the Java Dataset ($Accuracy^c$\%)}
\small
\label{tab:java_overall_perfomance}
\begin{threeparttable}
\setlength{\tabcolsep}{2.5pt}
\begin{tabular}{lccccccccccc}
\toprule
Method               & SB          & ES          & RJ          & GU          & RE          & DU          & GH          & ZX          & DR          & EB          & Overall          \\ \hline
\multicolumn{12}{c}{\cellcolor[HTML]{EFEFEF}\textit{Previous Untangling Approaches}}                                                                                           \\ \hline
Barnett et al.       & 21          & 27          & 24          & 25          & 19          & 18          & 24          & 20          & 18          & 21          & 17          \\
Herzig et al.        & 25          & 32          & 27          & 27          & 25          & 20          & 27          & 26          & 29          & 24          & 23          \\
SmartCommit          & 29          & 34          & 31          & 33          & 29          & 24          & 33          & 34          & 32          & 27          & 30          \\
Base-1               & 14          & 19          & 18          & 21          & 22          & 15          & 22          & 21          & 17          & 13          & 17          \\
Base-2               & 23          & 22          & 28          & 31          & 24          & 19          & 27          & 29          & 29          & 22          & 24          \\
UTango               & 32          & 36          & 33          & 37          & 34          & 30          & 35          & 38          & 34          & 31          & 34          \\
HD-GNN               & 37          & 40          & 38          & 42          & 36          & 32          & 40          & 42          & 39          & 36          & 38          \\ \hline
\multicolumn{12}{c}{\cellcolor[HTML]{EFEFEF}\textit{\textit{ColaUntangle} Variants}}                                                                                                       \\ \hline

$LLM_{ZeroShot}$                           &      59      &      66      &      57      &      47      &      47      &      57      &      51      &      65      &      64      &      55      &      59      \\
$LLM_{ZeroShotCoT}$                        &      61      &      69      &      60      &      66      &      52      &      60      &      48      &      63      &      64      &      56      &      63      \\
\textit{ColaUntangle}$_{no\_comments}$     & \textbf{68}  &      66      &   \textbf{69}  &      67      &   \textbf{60}  &      65   &     49 &      71             &      73     &      62      &   67  \\
\textit{ColaUntangle}                      &      66      &   \textbf{72}  &      67      &   \textbf{70}  &      53      &   \textbf{69}  &\textbf{54}  &   \textbf{72}  &\textbf{73}           &   \textbf{63}  &   \textbf{69}  \\
\bottomrule
\end{tabular}

\begin{tablenotes}
\footnotesize
\item Since previous studies do not report $Accuracy^a$\% on the Java dataset, we follow their practice and report only $Accuracy^c$\% to ensure a fair comparison.
\\
\textit{ColaUntangle}$_{no\_comments}$ refers to running \textit{ColaUntangle} on the dataset excluding comments.
\end{tablenotes}
\end{threeparttable}
\vspace{-0.3cm}
\end{table}

\subsection{RQ1: Overall Performance}
Table~\ref{tab:c_overall_perfomance} compares metrics on the C\# dataset (including nine GitHub projects: CL: Commandline, CM: CommonMark, HF: Hangfire, HU: Humanizer, LE: Lean, NA: Nancy, NJ: Newtonsoft.Json, NI: Ninject, RS: RestSharp). Compared to previous untangling approaches, \textit{ColaUntangle} outperforms SOTA, improving \textit{Barnett et al.}~\cite{barnett2015Heuristic-rule}, \textit{Herzig et al.}~\cite{herzig2016impact}, \textit{$\delta$-PDG+CV}~\cite{parțachi2020flexeme}, \textit{Flexeme}~\cite{parțachi2020flexeme}, \textit{Utango}~\cite{li2022utango} and \textit{HD-GNN}~\cite{fan2024HD-GNN} by 928.57\%, 157.14\%, 105.71\%, 111.76\%, 56.52\% and 44.00\% for $Accuracy^c$; 615.38\%, 40.91\%, 10.71\%, 12.05\%, 3.33\% and 3.33\% for $Accuracy^a$. For LLM-Based baselines, \textit{ColaUntangle} improves $LLM_{ZeroShot}$ and $LLM_{ZeroShotCoT}$ by 16.67\% and 16.39\% on $Accuracy^c$, and by 3.33\% on $Accuracy^a$ for both. Compared to \textit{ColaUntangle}$_{no\_comments}$, which excludes comment statements, \textit{ColaUntangle} achieves a 2.86\% higher $Accuracy^c$, while $Accuracy^a$ remains the same.

Table~\ref{tab:java_overall_perfomance} shows the Java dataset comparison (including ten GitHub projects: SB: spring-boot, ES: elasticsearch, RJ: RxJava, GU: guava, RE: retrofit, DU: dubbo, GH: ghidra, ZX: zxing, DR: druid, EB: EventBus), where \textit{ColaUntangle} improves \textit{Barnett et al.}~\cite{barnett2015Heuristic-rule}, \textit{Herzig et al.}~\cite{herzig2016impact}, \textit{Base-1}, \textit{Base-2}, \textit{SmartCommit}~\cite{shen2021smartcommit} \textit{Utango}~\cite{li2022utango} and \textit{HD-GNN}~\cite{fan2024HD-GNN} by 305.88\%, 200.00\%, 305.88\%, 187.50\%, 130.00\%, 102.94\% and 81.58\% for $Accuracy^c$. For LLM-based baselines, \textit{ColaUntangle} improves over $LLM_{ZeroShot}$ and $LLM_{ZeroShotCoT}$ by 16.95\% and 9.52\% respectively, and improves over \textit{ColaUntangle}$_{no\_comments}$ by 2.99\% for $Accuracy^c$.

The results show that \textit{ColaUntangle} consistently outperforms all baseline models, with notable improvements in the $Accuracy^c$ metric. Importantly, unlike most existing methods, \textit{ColaUntangle} can untangle comment statements, which are common cosmetic edits in real-world commits, making it more practical for software development. \textit{ColaUntangle}$_{no\_comments}$ also surpasses all prior approaches, indicating that while comments can slightly enhance performance, our framework remains effective even without them. To be note, the single LLM's overall accuracy outperforms commit untangling specific baselines, indicating the effectiveness of LLM in capturing the semantic information of tangled commit and executing the untangling task.

Regarding collaborative consultation rounds, the average number of rounds is 1.39: 75.44\% of cases converge within one round, 12.11\% within two rounds, and 13.55\% reach the maximum of three rounds. Notably, in 48.91\% of evaluated commits, the explicit and implicit worker agents initially produced non-consensual results, underscoring the importance of collaborative consultation for resolving divergent untangling perspectives.

\textcolor{black}{To further understand the nature of the evaluation datasets, we conducted a manual annotation study to classify the dependency types present in tangled commits. We randomly sampled 311 tangled commits from the C\# dataset and 375 from the Java dataset (95\% confidence level, 5\% margin of error). Each sampled commit was independently annotated by two experienced developers and classified as \textit{semantic-only} (containing only implicit dependencies), \textit{structure-driven} (containing only explicit dependencies), or \textit{hybrid} (containing both). We found the vast majority of tangled commits (93.25\% for C\# and 91.73\% for Java) are hybrid, simultaneously containing both explicit and implicit dependencies. This finding validates our design of employing both an explicit and an implicit worker agent, as the overwhelming majority of real-world tangled commits require reasoning over both dependency types simultaneously.}

\begin{figure}[htbp]
    \centering
    \setlength{\abovecaptionskip}{0.1cm}
    \begin{subfigure}[b]{0.37\textwidth}
        \centering
        \includegraphics[width=1.0\linewidth]{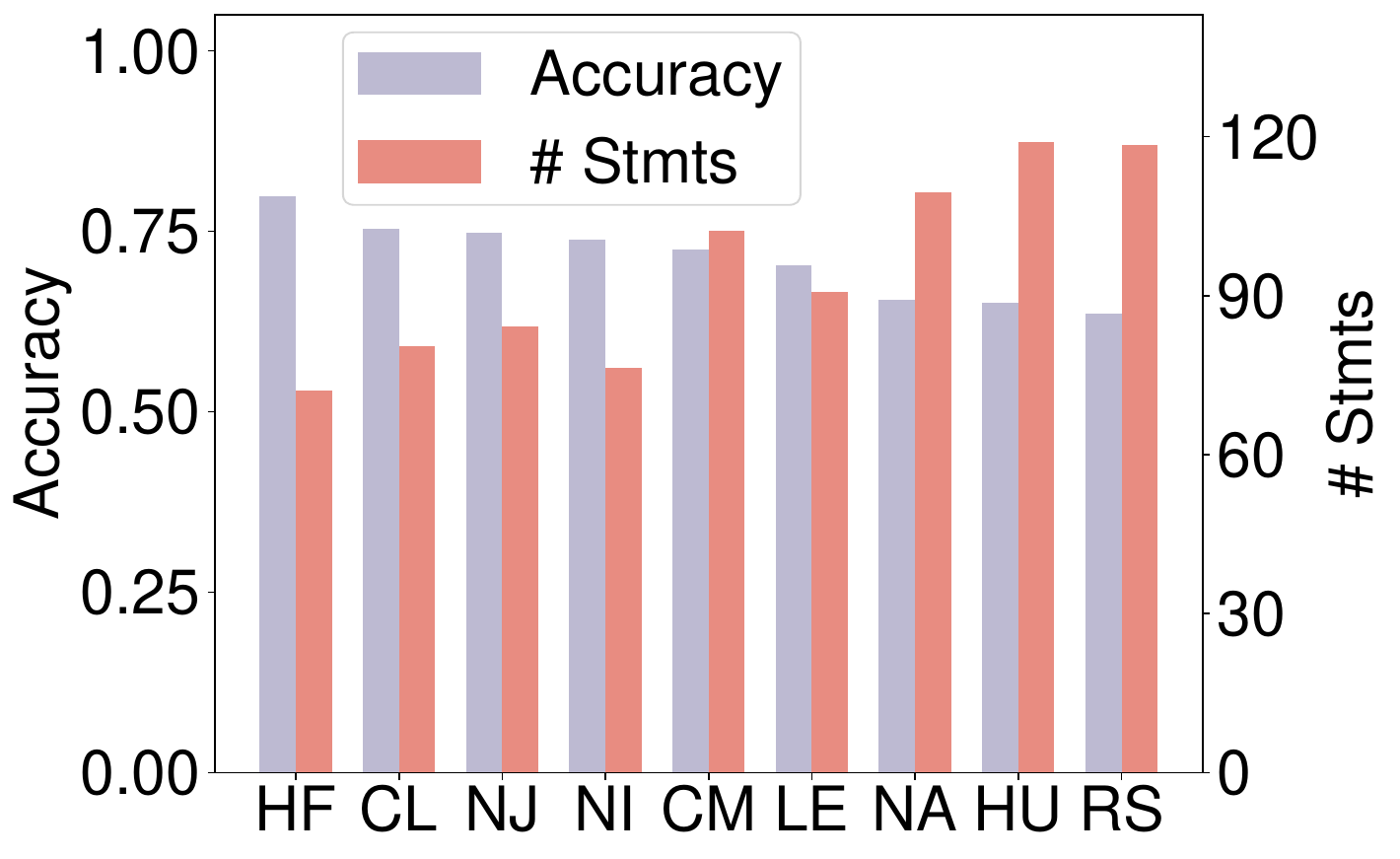}
        \caption{C\# Repositories}
        \label{fig:c_stmt}
    \end{subfigure}
    \begin{subfigure}[b]{0.45\textwidth}
        \centering
        \includegraphics[width=1.0\linewidth]{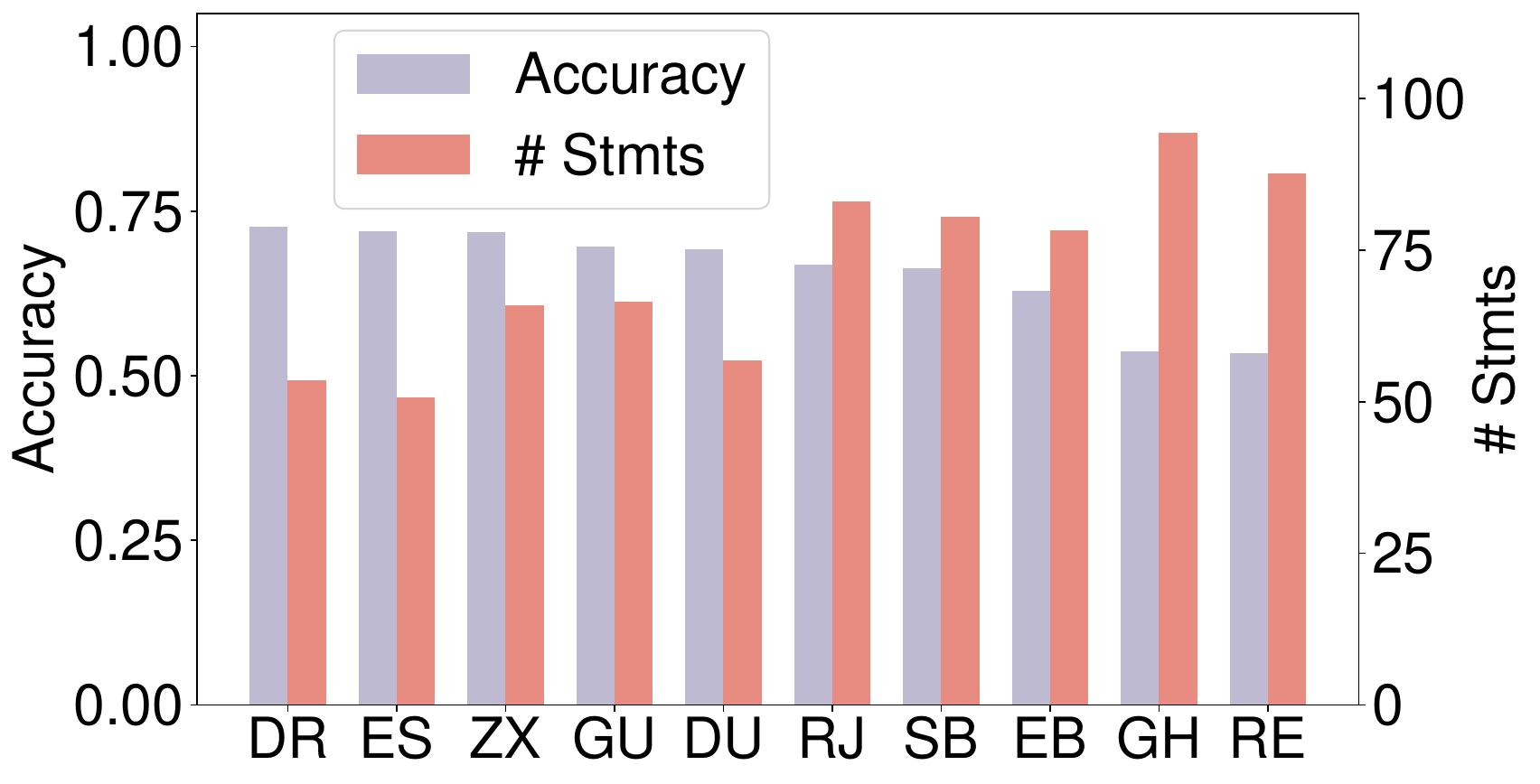}
        \caption{Java Repositories}
        \label{fig:c_acc}
    \end{subfigure}
    \caption{Average $Accuracy^c$ and Number of Changed Statements in Tangled Commits of Each Repository}
    \label{fig:repo_acc_stmt}
    \vspace{-0.3cm}
\end{figure}

We also investigate the factors contributing to accuracy differences across repositories. As shown in Figure \ref{fig:repo_acc_stmt}, for both the C\# and Java datasets, the purple bars represent the average $Accuracy^c$ in each repository, while the red bars indicate the average number of changed statements of tangled commits in each repository. We observe a general trend where $Accuracy^c$ decreases as the number of changed statements increases. This may be because abundant changes in the commit cause LLMs forgetting context or untangling commits more finely than humans expect. These findings suggest future work on hierarchical prompting or human-in-the-loop approaches to better align with human expectations.

\vspace{1mm}
\begin{custommdframed}
\textbf{\textit{Answer to RQ1:}} \textit{ColaUntangle} demonstrates improvements over prior baselines on both datasets, improving $Accuracy^c$ by 44\% on the C\# dataset and 82\% on the Java dataset over the best existing baseline. The results highlight its potential for practical application in untangling commits.
\end{custommdframed}
\vspace{0mm}

\subsection{RQ2: Ablation Study}
We conduct an ablation study to evaluate the contribution of each component within \textit{ColaUntangle}. As shown in Table \ref{tab:ablation_study}, we define four ablation settings:
\textcolor{black}{
(1) \textbf{w/o Explicit Contexts}: Removes explicit contexts ($G_{Explicit}$) extracted from the PDGs, retaining only implicit contexts.}
\textcolor{black}{
(2) \textbf{w/o Implicit Contexts}: Removes implicit contexts ($G_{Implicit}$) extracted from the PDGs, retaining only explicit contexts.}
\textcolor{black}{
(3) \textbf{w/o Explicit Contexts and Implicit Contexts}: Removes explicit and implicit contexts extracted from the PDGs.}
\textcolor{black}{
(4) \textbf{w/o Collaborative Consultation}: Uses a single LLM model to mine both explicit and implicit dependencies without the multi-agent consultation process.}
\textcolor{black}{
(5) \textbf{w/o Explicit Worker Agent (W1) and Collaborative Consultation}: Retains only the implicit worker agent to perform untangling. Since there is only a single agent remaining, the collaborative consultation process becomes unnecessary, and the reviewer agent is thus excluded.}
\textcolor{black}{
(6) \textbf{w/o Implicit Worker Agent (W2) and Collaborative Consultation}: The converse of (5), retaining only the explicit worker agent.}

\textcolor{black}{
From the results, we observe the following findings\footnote{For the sake of brevity, all mentions of performance degradation in this section refer specifically to a drop in $Accuracy^c$, as it better reflects the model's performance in real-world tasks compared to $Accuracy^a$.}:
}

\begin{table}[h!t]
\setlength{\abovecaptionskip}{0.1cm}
\caption{Ablation Study Results for Each Module ($Accuracy^c$\%/$Accuracy^a$\%)}
\label{tab:ablation_study}
\begin{threeparttable}
\small
{\color{black}
\begin{tabular}{lccc}
\toprule
Method                    & C\# Dataset    & Java Dataset   & Overall        \\ \hline
w/o Explicit Contexts     & 69/92          & 67/92          & 69/92          \\
w/o Implicit Contexts     & 67/92          & 65/91          & 66/91          \\
w/o Explicit \& Implicit Contexts & 64/91  & 65/91          & 65/91          \\
w/o Collaborative         & 60/90          & 59/91          & 59/91          \\
w/o W1 \& w/o Collaborative  & 57/90          & 54/90          & 56/90        \\  
w/o W2 \& w/o Collaborative  & 58/90          & 55/92          & 57/91      
         \\ \hline
\textit{ColaUntangle}                 & \textbf{72/93} & \textbf{69/93} & \textbf{71/93} \\ 
\bottomrule
\end{tabular}
}
\begin{tablenotes}
\footnotesize
\item W1(2) refer to worker agent that expert in discovering explicit (implicit) dependency between code changes.
\end{tablenotes}
\end{threeparttable}
\vspace{-0.3cm}
\end{table}

\textcolor{black}{
Removing explicit contexts alone (setting 1) results in an $Accuracy^c$ drop from 71\% to 69\% (2.82\%), while removing implicit contexts alone (setting 2) causes a larger drop from 71\% to 66\% (7.04\%). Removing both contexts (setting 3) leads to the largest context-related drop to 65\% (8.45\% degradation). This confirms that both explicit and implicit contexts provide valuable information for commit untangling. Notably, removing implicit contexts causes a substantially larger degradation than removing explicit contexts, suggesting that while LLMs can partially infer structural dependencies from code diffs alone, implicit dependencies—such as semantic similarity and shared development purpose—are harder to detect without the richer contextual information provided by $G_{Implicit}$.
}

\textcolor{black}{
Excluding Collaborative Consultation (setting 4) leads to the most significant performance degradation (16.90\%), highlighting its critical role in achieving optimal untangling results. Removing either the explicit or implicit worker agent (settings 5 and 6) also results in significant performance drops (18.99\% and 19.72\%, respectively). However, it is important to note that these two settings inherently exclude Collaborative Consultation due to the presence of only a single agent, making their performance degradation partly attributable to the absence of collaborative consultation. When we isolate the effect by excluding the performance drop caused by the lack of collaborative consultation, settings 5 and 6 show additional reductions of only 2.09\% and 2.82\%, respectively. This indicates that while incorporating both explicit and implicit worker agents enhances performance, the collaborative consultation mechanism itself contributes the most substantial improvement in our framework. Removing both explicit contexts and implicit contexts (setting 3) results in a performance decrease of 8.45\%, suggesting that the structured context improves our model's untangling effectiveness. However, compared to the substantial improvements brought by collaborative consultation, Info Tools mainly serves as foundational support by providing essential contextual information rather than driving the core reasoning and decision-making process.
}

\vspace{1mm}
\begin{custommdframed}
\textbf{\textit{Answer to RQ2:}} The ablation results suggest that while structured context information and the integration of both explicit and implicit worker agents supports untangling accuracy, the collaborative consultation mechanism plays a more critical role in achieving the effectiveness of \textit{ColaUntangle}.
\end{custommdframed}
\vspace{0mm}

\textcolor{black}{\subsection{RQ3: Sensitivity Analysis}}

\textcolor{black}{
In this section, we investigate the sensitivity of \textit{ColaUntangle} to two key design choices: the selection of the underlying LLM and the hop count used to construct the implicit context graph, which directly influence the quality of the information available to each agent and the cost of reaching consensus.
}

\textcolor{black}{
\subsubsection{Sensitivity to Different LLMs}
To evaluate the generalization capability of our framework, we include both closed-source proprietary models (GPT-4o, GPT-4o-mini from OpenAI, Claude-4-sonnet from Anthropic) and open-source frontier models (DeepSeek-V3 and Qwen3-235b) in our comparison. We compare \textit{ColaUntangle} instantiated with these five models respectively. The performance presented in Table \ref{tab:performance_llms} shows that all these LLMs can effectively untangle commits, with DeepSeek-V3 performing slightly better. This suggests that while \textit{ColaUntangle} is generally robust to the underlying LLM choice, stronger reasoning models still yield marginal performance gains. Furthermore, we analyze the average number of collaborative rounds required for consensus: DeepSeek-V3 (1.39), GPT-4o-mini (1.65), GPT-4o (1.34), Qwen3-235b (1.41), and Claude-4-sonnet (1.73). Notably, DeepSeek-V3 not only attains leading accuracy but also requires fewer consensus rounds compared to most competing models (except GPT-4o), highlighting its efficiency in balancing performance and computational cost.}

\begin{table}[h!t]
\setlength{\abovecaptionskip}{0.1cm}
\caption{Performance of Different LLMs ($Accuracy^c$\%/$Accuracy^a$\%)}
\label{tab:performance_llms}
\small
\begin{threeparttable}
\begin{tabular}{lccc}
\toprule
Method                & C\# Dataset    & Java Dataset   & Overall        \\ \hline
GPT-4o-mini            & \textbf{72/93} & 64/91          & 69/93          \\
GPT-4o                 & 68/92          & 67/93          & 67/93          \\
Qwen3-235b             & 71/93          & 66/92          & 69/92          \\
Claude-4-sonnet        & 70/93          & 68/92          & 69/93          \\ \hline
ColaUntangle (DeepSeek-V3) & \textbf{72/93} & \textbf{69/93} & \textbf{71/93}  \\ 
\bottomrule
\end{tabular}

\end{threeparttable}
\vspace{-0.3cm}
\end{table}

\textcolor{black}{
\subsubsection{Sensitivity to Hop Count in Implicit Context}
The implicit context ($G_{Implicit}$) is constructed by extracting changed nodes and their neighbors within a certain hop distance in $G_{Merged}$. To investigate the impact of hop count on untangling performance, we evaluate \textit{ColaUntangle} with 1-hop, 2-hop, and 3-hop implicit contexts. As shown in Table~\ref{tab:hop_count}, increasing the hop count from 1 to 2 yields a marginal improvement in $Accuracy^c$ (from 71\% to 72\%), while further increasing to 3 hops leads to a slight decrease (70\%). Meanwhile, the average number of nodes in the implicit context grows substantially from 235 (1-hop) to 322 (2-hop) and 405 (3-hop), resulting in significantly higher LLM token consumption and computational cost.}

\begin{table}[h]
\setlength{\abovecaptionskip}{0.1cm}
\caption{\textcolor{black}{Impact of Hop Count on Implicit Context}}
\label{tab:hop_count}
\centering
\small
\color{black}{
\begin{tabular}{lccc}
\toprule
                              & 1-hop          & 2-hop          & 3-hop          \\ \hline
Overall $Accuracy^c$/$Accuracy^a$ (\%) & 71/93          & 72/93          & 70/92          \\
Avg. Node Count in $G_{Implicit}$     & 235            & 322            & 405            \\
\bottomrule
\end{tabular}
}
\vspace{-0.2cm}
\end{table}

\textcolor{black}{
These results suggest that 1-hop neighbors provide a well-balanced trade-off between contextual richness and computational efficiency. The 2-hop setting offers only a negligible accuracy gain (+1\%) at the cost of a 37\% increase in average node count, while the 3-hop setting introduces excessive noise from distant, less relevant nodes, leading to both degraded accuracy and a 72\% increase in node count. We therefore adopt 1-hop as the default setting for \textit{ColaUntangle}.
}

\vspace{1mm}
\begin{custommdframed}

\textcolor{black}{\textbf{\textit{Answer to RQ3:}} Our evaluation demonstrates that \textit{ColaUntangle} is robust to the choice of underlying LLM, with DeepSeek-V3 achieving the optimal balance of accuracy and efficiency. Regarding the hop count in implicit context construction, 1-hop provides the best trade-off between contextual richness and computational cost.}
\end{custommdframed}
\vspace{0mm}

\section{Discussion}

\textcolor{black}{
This section presents a multi-faceted analysis of \textit{ColaUntangle} beyond quantitative results. We first examine misclassified cases to identify recurring failure patterns and inform future improvements. We then conduct a qualitative comparison on a representative correct case to illustrate how multi-agent collaboration contributes to untangling accuracy. Next, we report a user study to assess whether the generated explanations are interpretable and useful to practitioners. Finally, we perform a data leakage analysis to verify that the observed performance stems from genuine reasoning capability rather than memorization of training data.
}

\begin{table}[h!t]
\setlength{\abovecaptionskip}{0.1cm}
\caption{Type of Error Cases}
\label{tab:error_case_type}
\begin{threeparttable}
\footnotesize
\begin{tabular}{llc}
\toprule
Type                                         & Sub-Type               & Number \\\hline
\multirow{3}{*}{Fine Grouping Granularity}   & Judgment Error & 14     \\
                                             & Composite Commit       & 17     \\
                                             & File Formatting        & 3      \\ \hline
\multirow{2}{*}{Coarse Grouping Granularity} & Similar Change         & 11     \\
                                             & Mergeable Logic        & 5      \\ 
\bottomrule
\end{tabular}
\end{threeparttable}
\vspace{-0.1cm}
\end{table}

\subsection{Error Cases}

In the RQ1 experiments, although \textit{ColaUntangle} performed well overall, it still misclassified some cases. To support future improvements, we conducted a manual analysis following Fan et al.~\cite{fan2024HD-GNN}, randomly sampling 50 misclassified cases. As shown in Table \ref{tab:error_case_type}, we grouped the errors into two main types: fine and coarse grouping granularity. Fine grouping granularity refers to \textit{ColaUntangle} untangles a tangled commit into concerns at finer granularity than the standard answer. In contrast, coarse grouping granularity refers to \textit{ColaUntangle} untangles a tangled commit into concerns at coarser granularity than the standard answer. Further analysis identified five sub-types that specify the scenarios of these errors.

\begin{figure*}
    \centering
    \setlength{\abovecaptionskip}{0.1cm}
    \includegraphics[width=0.85\linewidth]{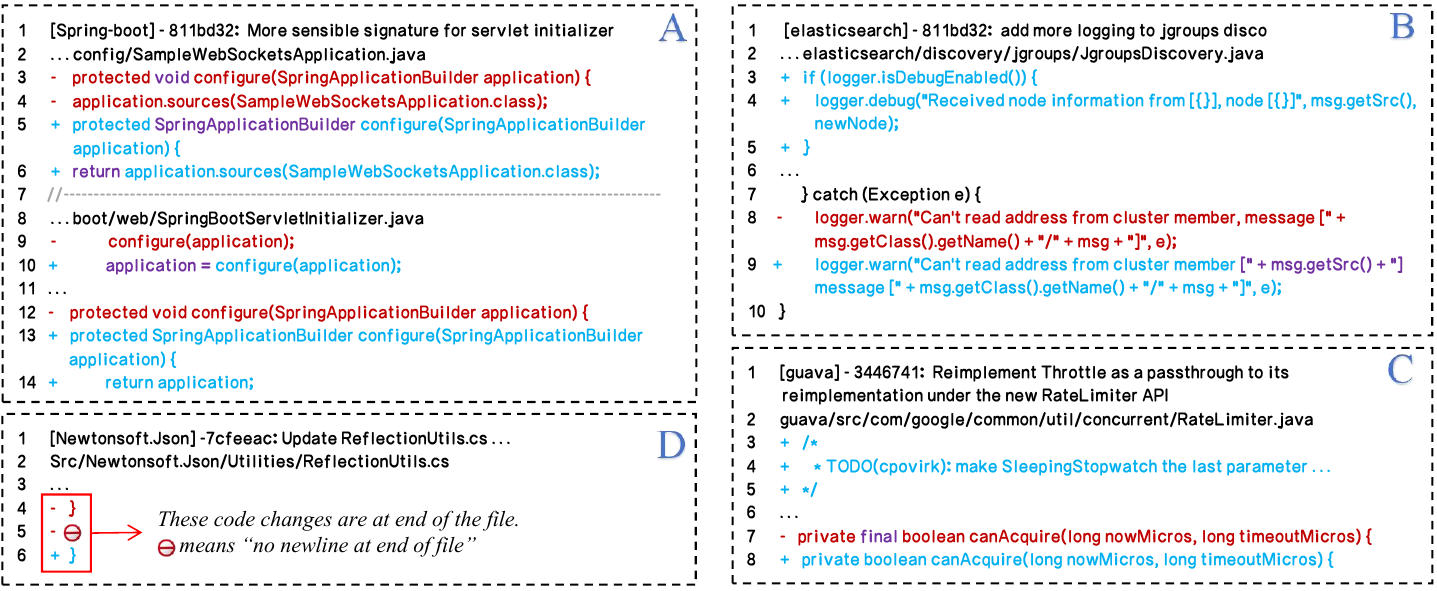}
    \caption{Examples of Error Cases Related to Fine Grouping Granularity}
    \label{fig:error_case_fine}
    \vspace{-0.1cm}
\end{figure*}

For \textbf{fine grouping granularity}, we observed three representative sub-types: \textit{judgment errors} (14 cases), \textit{composite commit} (17 cases), and \textit{file formatting} (3 cases). \textit{Judgment errors} (14 cases) occur when \textit{ColaUntangle} misinterprets the relationship between changes. For example, in Figure \ref{fig:error_case_fine} A, changes across two files belong to a single concern related to signature revision, but \textit{ColaUntangle} misclassified them into two concerns by treating lines 9–10 as an isolated function call modification (\texttt{configure}) unrelated to the other file. Similarly, in Figure \ref{fig:error_case_fine} B, changes at lines 3–5 and 8–9 belong to the same concern (\textit{add more logging to jgroups disco}), but \textit{ColaUntangle} separated them because it focused on different logger methods (\texttt{debug} vs. \texttt{warn}). 
\textit{Composite commit} (17 cases) occur when the standard atom commit in the dataset actually contains multiple unrelated activities. For instance, Figure \ref{fig:error_case_fine} C shows a commit that modifies comments for the \texttt{create} function (lines 3–5) and changes the signature of \texttt{canAcquire}, with the two code changes separated by over 240 lines and lacking any logical connection. \textit{ColaUntangle} classified them into separate concerns, diverging from the dataset’s single-concern annotation. 
\textit{File formatting} (3 cases) involve changes unrelated to human edits, such as automatic formatting. As shown in Figure \ref{fig:error_case_fine} D, the first change in this commit is related to variable revise (not shown in Figure \ref{fig:error_case_fine} D), the second change adds a newline at the end of the file, which is required by POSIX standards but inconsistently enforced by code editors \cite{posixlinedstand}. Although such changes have no functional impact and are often ignored by developers, \textit{ColaUntangle} classified them as a separate formatting concern, resulting in misclassification.

\begin{figure*}
    \centering
    \setlength{\abovecaptionskip}{0.1cm}
    \includegraphics[width=0.9\linewidth]{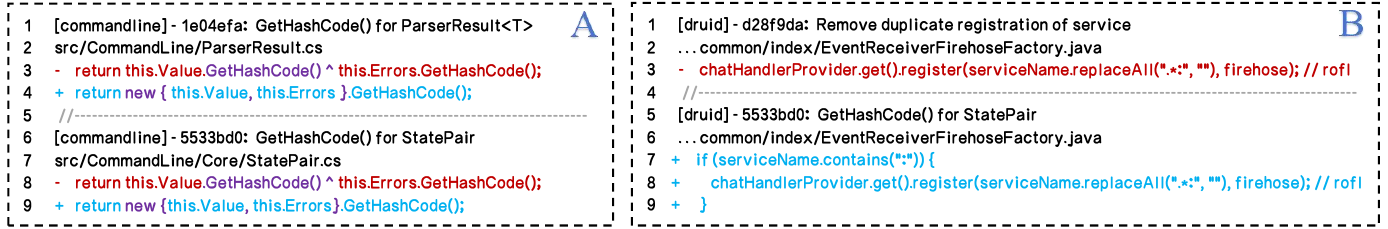}
    \caption{Examples of Error Cases Related to Coarse Grouping Granularity}
    \label{fig:error_case_coarse}
    \vspace{-0.1cm}
\end{figure*}
Since the datasets are built by selecting atomic commits and artificially tangling them, some standard atomic commits may exhibit high similarity in content or logic. This can cause \textit{ColaUntangle} to misclassify them at a \textbf{coarse grouping granularity}. Specifically, we observed two common sub-types: \textit{similar change} (11 cases) and \textit{mergeable logic} (5 cases). 
\textit{Similar change} refers to cases where two atomic commits contain highly similar modifications. As shown in Figure~\ref{fig:error_case_coarse}A, both commits modify different files with nearly identical changes to the return value of \texttt{GetHashCode}, leading \textit{ColaUntangle} to group them as one concern.
\textit{Mergeable logic} occurs when changes in two atomic commits are tightly related. As shown in Figure~\ref{fig:error_case_coarse}B, one commit removes a registration statement, and the other re-adds it within an \texttt{if} block at the same location (line 47). \textit{ColaUntangle} interpreted them as a single concern introducing conditional registration.

Overall, our approach occasionally untangles commits at an unexpected granularity, revealing challenges in both concern boundary modeling and dataset construction. In future work, we plan to improve concern modeling with human-in-the-loop interaction and develop more reliable methods for building commit untangling datasets to support this line of research. \textcolor{black}{Additionally, we plan to investigate the incorporation of commit messages as supplementary semantic context for post-hoc untangling scenarios, where commit messages are already available and could provide valuable cues about developer intent.}


\subsection{Correct Cases}

To further validate the effectiveness of our approach, we randomly selected a representative case from the dataset and conducted a comparative evaluation of ColaUntangle, UTango~\cite{li2022utango}, and a single LLM. We selected UTango as a key comparator because it partially shares common structural data sources (utilizing $\delta$-PDG, which informs our structured code change) and its methodology touches upon the concepts of explicit and implicit factors. Specifically, UTango employs a sequential, two-stage process: extracting feature embeddings from the $\delta$-PDG for initial clustering (explicit context), followed by a code-clone detection process to refine groupings based on similarity (implicit dependency). This sequential approach is a critical point of contrast with our collaborative mechanism. We selected the single LLM (DeepSeek-V3) as a second comparator to isolate and intuitively observe the performance of LLM on direct reasoning. This baseline operates without access to the structured explicit/implicit contexts or the rule-based guidance provided to our multi-agent framework.
Through this comparative setting, we aim to highlight the contribution of ColaUntangle’s multi-agent reasoning and context-aware mechanisms to improving untangling accuracy and interpretability.

\begin{figure}[H]
    \centering
    \includegraphics[width=0.85\linewidth]{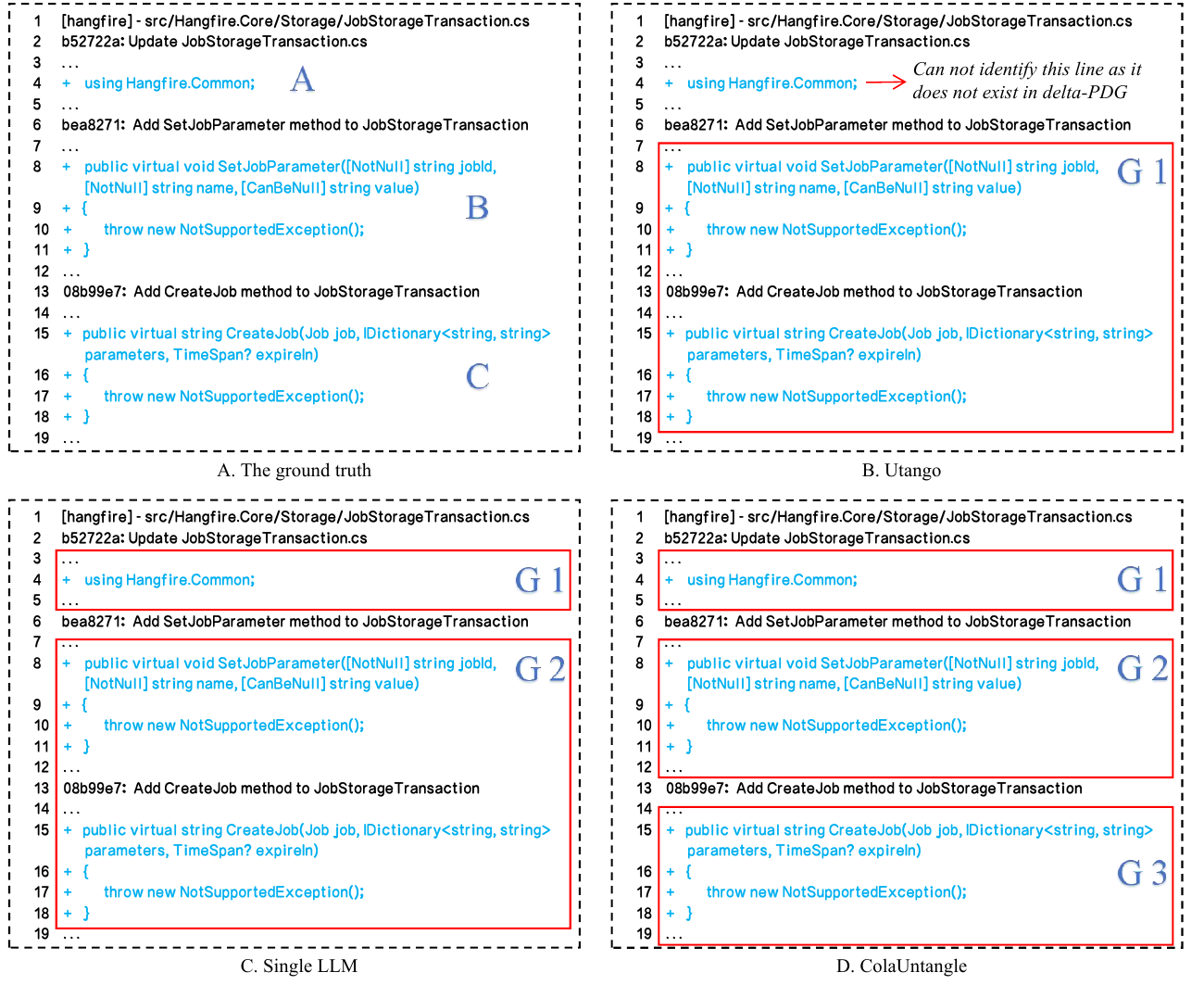}
    \caption{Untangling Result of Different Model}
    \label{fig:compare}
\end{figure}

As shown in Figure~\ref{fig:compare}, the commit contained three distinct changes: a file update, and the addition of two new functions, \texttt{SetJobParameter} and \texttt{CreateJob}. While UTango and the single LLM failed to untangle this commit correctly, our framework successfully generated the correct answer, which is not a trivial task due to the subtle dependencies between the code changes. UTango failed for two key reasons: it could not identify the first code change because the $\delta$-PDG representation was incomplete (lacking the import statement at the file's beginning), and it incorrectly grouped the two new functions together based on a structural similarity detected by its code clone detection process, ignoring their functional independence. Similarly, the single LLM also grouped the two functions together, explaining its decision based on a generic principle of separating ``cosmetic edits from functional changes," but it failed to discern the distinct purposes of the two new functions.

\begin{figure}
    \centering
    \includegraphics[width=0.85\linewidth]{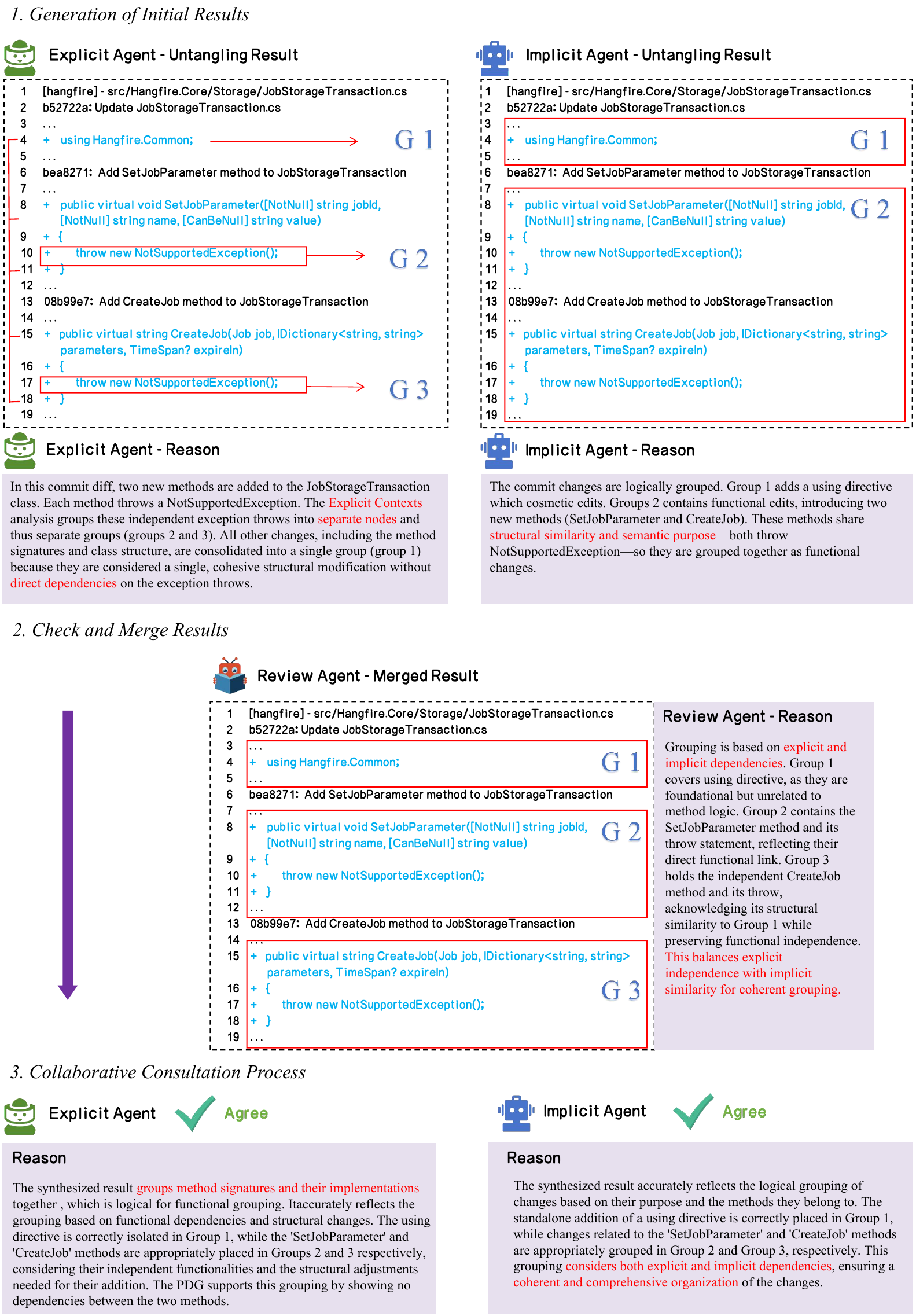}
    \caption{Untangling Process of ColaUntangle}
    \label{fig:process}
\end{figure}

The effectiveness of \textit{ColaUntangle} lies in its underlying process, as shown in our step-by-step analysis (Figure~\ref{fig:process}). The Explicit Agent, using the explicit contexts and original commit diff, correctly identified the two new functions as separate entities, while the Implicit Agent's semantic analysis noted their structural and semantic similarities. The Reviewer Agent then synthesized and balanced these two viewpoints, ultimately generating a correct grouping that preserved the functional independence of the new functions. The collaborative consultation process then led all agents to agree on this correct grouping. This example illustrates the combined power of our key ideas: the LLM's superior semantic understanding, its ability to deal with cosmetic edits and other non-functional changes, and its capacity to use the explicit and implicit contexts as an active component to infer subtle dependencies. The collaborative framework ensures a balanced and accurate result by reconciling the distinct explicit and implicit viewpoints, which together make ColaUntangle's performance superior to that of prior work and single-model baselines.

\vspace{0.3cm}

\textcolor{black}{
\subsection{User Study}}

\textcolor{black}{
Beyond quantitative accuracy, we conducted a user study to evaluate whether the explanations generated by \textit{ColaUntangle} are genuinely useful and interpretable to practitioners. We recruited 4 participants with over 5 years of software development experience. We randomly sampled 25 tangled commits from the evaluation datasets and collected the untangling results along with their explanations generated by \textit{ColaUntangle}. Each participant independently evaluated 10 commits (5 shared across all participants to enable inter-rater agreement analysis, and 5 unique per participant). The explanations and related contextual information (e.g., commit messages, source code diffs and untangle process) were in randomized order to avoid bias. Three dimensions were assessed using a 5-point Likert scale: \textbf{Coherence} (logical clarity and consistency of the explanation), \textbf{Usefulness} (whether the explanation helps understand the grouping rationale), and \textbf{Alignment} (whether the reasoning matches how a human developer would approach the untangling task)~\cite{doshi2017towards, miller2019explanation, hoffman2018metrics}. 
}

\begin{figure}
    \centering
    \includegraphics[width=0.85\linewidth]{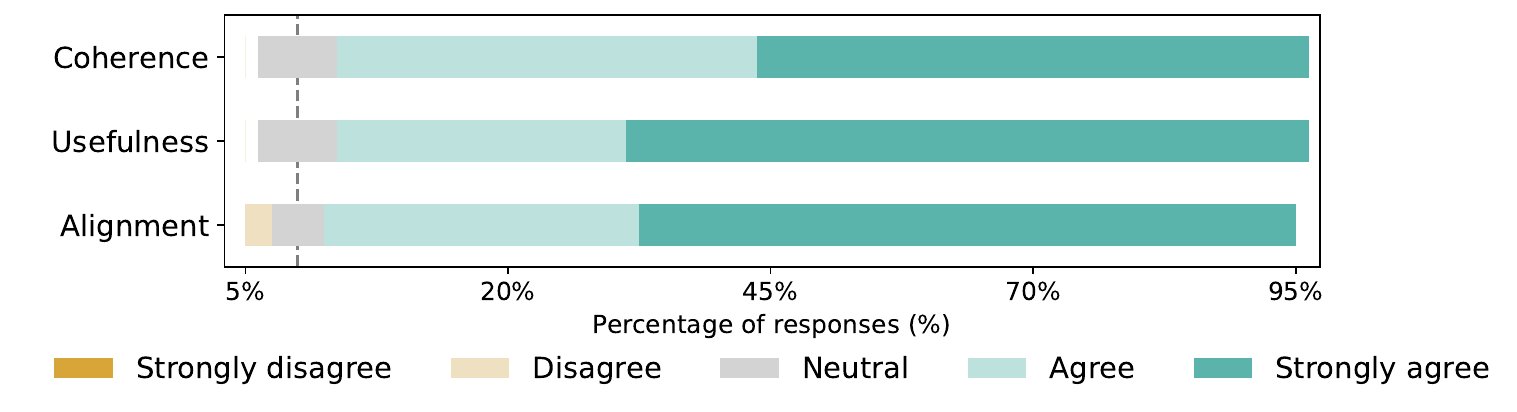}
    \caption{\textcolor{black}{Evaluation Results of User Study}}
    \label{fig:user_study}
\end{figure}

\textcolor{black}{
The results are shown in Fig.~\ref{fig:user_study}, across 40 ratings (4 participants $\times$ 10 commits each), the explanations received high scores on all three dimensions, with means of 4.45, 4.58, and 4.53 for Coherence, Usefulness, and Alignment respectively (overall mean = 4.52 out of 5). Substantial inter-rater agreement was observed among the five shared commits evaluated by all participants ($\kappa = 0.62$)~\cite{landis1977measurement}. Usefulness was rated highest, suggesting that the agent-generated explanations provide actionable insight into why specific file changes were grouped together—a property that purely predictive models lack. 
}

\textcolor{black}{
These findings indicate that the multi-agent architecture of \textit{ColaUntangle} not only improves untangling accuracy but also produces human-comprehensible justifications, addressing a key limitation of prior black-box approaches to commit untangling. We believe this interpretability property is particularly valuable in practice, as developers are more likely to trust and act on tool recommendations when the underlying reasoning is transparent and consistent with their own mental models of code change semantics.
}

\textcolor{black}{
We noticed that in one case the Alignment score dropped to 2, representing the most notable outlier in our study. In this case, \textit{ColaUntangle} correctly untangled a commit into three groups, matching the ground-truth atomic commits exactly: a bug fix that corrected the type-checking logic for the \texttt{Guid} variant, a version bump, and a set of new test cases specifically added to validate the \texttt{Guid} handling fix. Although this three-way split faithfully reflects the boundaries of the original atomic commits, one participant assigned a low Alignment score because, from a developer's perspective, the bug fix and its corresponding test cases intuitively belong to the same logical concern—fixing an issue and verifying it are two sides of the same task. The explanation justified the separation on structural grounds (production code versus test code), which is consistent with dependency-based grouping criteria but diverges from the intent-driven mental model the participant hold.
}

\textcolor{black}{
This discrepancy points to a fundamental challenge: the boundary of a concern is inherently subjective and context-dependent. Defining concern boundaries that align with human intuition remains an open problem, and we believe it warrants future investigation through approaches such as reinforcement learning from human feedback, where developer judgments can be incorporated as reward signals to guide the grouping strategy.
}

\textcolor{black}{
\subsection{Data Leakage Analysis}}

\textcolor{black}{
\label{subsec:data_leakage}
Since LLMs are trained on internet-scale data, a natural concern is whether the model has memorized parts of the evaluation datasets, leading to inflated performance. To investigate this, we constructed a post-cutoff benchmark by mining 300 tangled commits created after the release date of DeepSeek-V3 (December 26, 2024), following the same construction methodology as the original benchmarks~\cite{parțachi2020flexeme}. This ensures that these commits could not have appeared in the model's pretraining data. As shown in Table~\ref{tab:data_leakage}, the results of \textit{ColaUntangle} on this benchmark ($Accuracy^c$: 70\% for C\# and 66\% for Java) are comparable to those on the original datasets (72\% and 69\%). The LLM baselines similarly show no performance inflation on the post-cutoff benchmark (57\% vs.\ 60\% for C\#; 55\% vs.\ 59\% for Java). These results confirm that the performance of \textit{ColaUntangle} stems from its reasoning capabilities rather than memorization.
}

\begin{table}[h]
\setlength{\abovecaptionskip}{0.1cm}
\caption{\textcolor{black}{Results on Post-cutoff Benchmark vs.\ Original Datasets}}
\label{tab:data_leakage}
\centering
\footnotesize
\color{black}
\begin{tabular}{llcc}
\toprule
Dataset & Method & Post-cutoff $Accuracy^c$ (\%) & Original $Accuracy^c$ (\%) \\ \hline
\multirow{3}{*}{C\#} & $LLM_{ZeroShot}$ & 57 & 60 \\
 & \textit{ColaUntangle} & 70 & 72 \\
\hline
\multirow{3}{*}{Java} & $LLM_{ZeroShot}$ & 55 & 59 \\
 & \textit{ColaUntangle} & 66 & 69 \\
\bottomrule
\end{tabular}
\vspace{-0.3cm}
\end{table}

\section{Threats to Validity} 
\textbf{Internal validity.} 
One potential threat to internal validity arises from our definitions of explicit and implicit dependencies. Our framework categorizes relationships as either explicit or implicit, and while explicit rules are obvious (data and control dependency), it's impossible to manually enumerate every complex, context-dependent implicit semantic relationship that drives developer intent. To mitigate this, we shift our approach from a brittle, rule-based system to a flexible, reasoning-based paradigm. We acknowledge that manually summarized rules cannot cover all implicit cases. Instead, we perform extensive analysis of literature and code to define patterns covering the majority of critical implicit dependencies. We then provide our LLM-driven worker agents with both structured rules and rich contextual information derived from PDGs. Leveraging the LLM's semantic understanding and reasoning capabilities, the implicit worker agent is able to infer latent dependencies that extend beyond these manually formalized rules. The subsequent collaborative consultation mechanism, where the reviewer agent reconciles the distinct views of the explicit and implicit workers, ensures a comprehensive analysis over diverse dependency patterns. The strong performance of our method suggests that our implicit dependency coverage, combined with the LLM's inference power, is sufficient for robust practical performance.
Another threat arises from the reliance of PDGs on static analysis tools to model code. The accuracy and reliability of these tools can be limited, especially when handling complex language features, potentially leading to inaccurate analysis results. 

\textbf{External validity.}
A threat to external validity is potential data leakage, as the LLM may have been pre-trained on some open-source repositories used in our evaluation. However, this risk is minimal. The datasets are constructed by merging real atomic commits into new tangled commits, which are unlikely to appear in pretraining data. Moreover, the standalone LLM achieves only 60\% and 59\% $Accuracy^{c}$ on the C\# and Java datasets, while our method improves this to 72\% and 69\%, suggesting the gain comes from our approach. Additionally, we attempted to prompt the LLM to reproduce exact commit histories but failed, indicating that the model does not memorize verbatim and relies on reasoning rather than recall. \textcolor{black}{We further address this concern with a dedicated post-cutoff benchmark analysis in Section~\ref{subsec:data_leakage}.}
Another threat stems from the evaluation datasets themselves, as the tangled commits are constructed artificially by combining atomic commits. Although such datasets may include imprecise or non-representative data points, they are widely adopted in prior studies~\cite{parțachi2020flexeme, li2022utango, fan2024HD-GNN, xu2025COMUNT}. In the future, we plan to develop more realistic and precise datasets to better evaluate commit untangling approaches.

\section{Conclusion}
 
In this paper, we propose \textit{ColaUntangle}, the first LLM-driven collaborative model for commit untangling that jointly considers explicit and implicit dependencies among code changes. By integrating structured code change information with specialized agents in an iterative consultation process, \textit{ColaUntangle} enables deeper semantic reasoning and interpretable decision-making.
\textit{ColaUntangle} achieves substantial performance gains, improving $Accuracy^{c}$ by 44\% on the C\# dataset and 82\% on the Java dataset over the best-performing baseline. Our work introduces a new paradigm that simulates human-like collaborative reasoning, advancing the accuracy and transparency of automated commit untangling. Our tool and data are available at \url{https://github.com/ColaUntangle/ColaUntangle}.

\begin{acks}
This work is supported by the National Natural Science Foundation of China (Grant Nos. 62572030 and 92582204), the Fundamental Research Funds for the Central Universities, and the exploratory elective projects of the State Key Laboratory of Complex and Critical Software Environments. 

\end{acks}

\bibliographystyle{ACM-Reference-Format}
\bibliography{sample-base}

\end{document}